\documentclass[oneside,11pt]{article} % For LaTeX2e
\renewenvironment{abstract}
  {{\centering\large\bfseries Abstract\par}\vspace{0.7ex}%
    \bgroup
       \leftskip 20pt\rightskip 20pt\small\noindent\ignorespaces}%
  {\par\egroup\vskip 0.25ex}

\newenvironment{keywords}
{\vspace{0.05in}\bgroup\leftskip 20pt\rightskip 20pt \small\noindent{\bfseries
Keywords:} \ignorespaces}%
{\par\egroup\vskip 0.25ex}

\usepackage{times}
\usepackage{bm}
\usepackage{amsfonts}
\usepackage{graphicx}
\usepackage{algorithm,algorithmic}
\usepackage{amsfonts,amssymb,amsmath,amsthm} %sometimes replace mathtools with amsmath
\usepackage{xcolor}
\usepackage{url}
\usepackage{multirow}
\usepackage{booktabs}
\usepackage{enumitem}
\usepackage{subfig}
\usepackage[colorlinks,
            linkcolor=blue,
            citecolor=blue,
            urlcolor=magenta,
            linktocpage,
            plainpages=false]{hyperref}
\numberwithin{equation}{section}

\theoremstyle{plain}

\theoremstyle{definition}

\newcommand{\w}{\mathbf{w}}
\graphicspath{{fig/}}

\title{Online Feature Selection with Group Structure Analysis}
\author{
{\bf Jing Wang, Meng Wang, Peipei Li}\\
{\bf Zhongqiu Zhao, Xuegang Hu, Xindong Wu}\\
Department of Computer Science \\
Hefei University of Technology\\
Hefei, Anhui 230009, China\\
\and
{\bf Luoqi Liu}\\
Department of Electrical and Engineering\\
National University of Singapore\\
117583, Singapore\\
}
\date{}
\begin{document}
\maketitle

\begin{abstract}
Online selection of dynamic features has attracted intensive interest in recent years. However, existing online feature selection methods evaluate features individually and ignore the underlying structure of feature stream. For instance, in image analysis, features are generated in groups which represent color, texture and other visual information. Simply breaking the group structure in feature selection may degrade performance. Motivated by this fact, we formulate the problem as an online group feature selection. The problem assumes that features are generated individually but there are group structure in the feature stream. To the best of our knowledge, this is the first time that the correlation among feature stream has been considered in the online feature selection process. To solve this problem, we develop a novel online group feature selection method named OGFS. Our proposed approach consists of two stages: online intra-group selection and online inter-group selection. In the intra-group selection, we design a criterion based on spectral analysis to select discriminative features in each group. In the inter-group selection, we utilize a linear regression model to select an optimal subset. This two-stage procedure continues until there are no more features arriving or some predefined stopping conditions are met. %Our method has been applied
Finally, we apply our method to multiple tasks including image classification %, face verification
and face verification. Extensive empirical studies performed on real-world and benchmark data sets demonstrate that our method outperforms other state-of-the-art online feature selection %method
methods.
\end{abstract}

\begin{keywords}
Online feature selection, Streaming feature, Group structure, Classification, Face Verification.
\end{keywords}

\section{Introduction}
\begin{figure}
\centering
   \subfloat[An Image]{
        \includegraphics[width=0.230\textwidth]{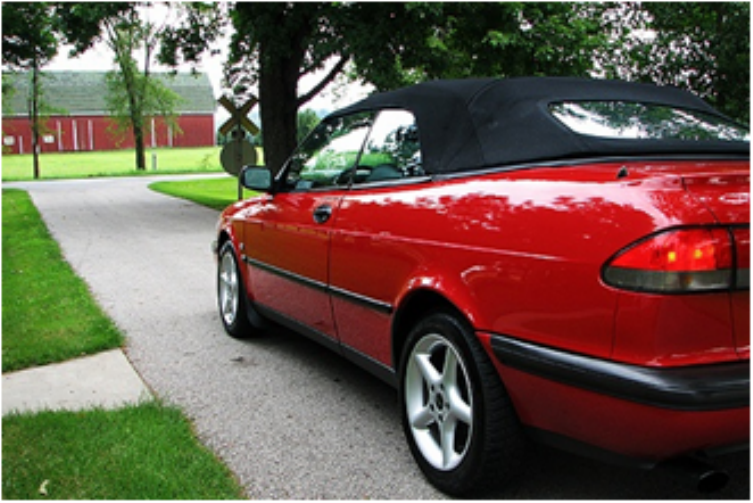}}
     \subfloat[HOG]{
        \includegraphics[width=0.230\textwidth]{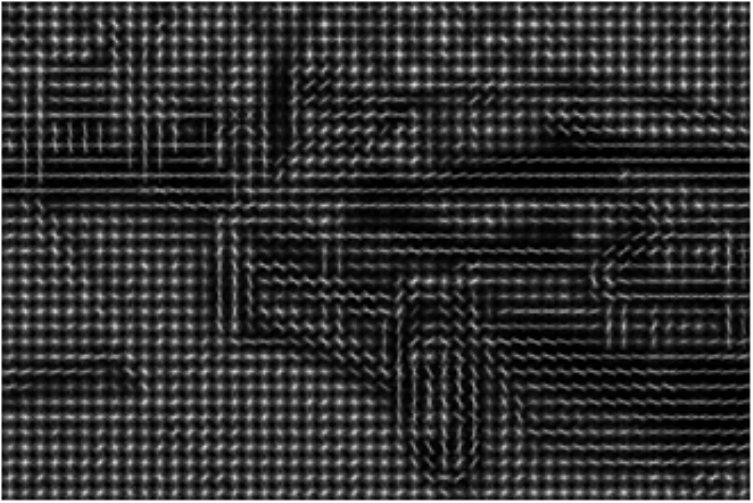}}
%        \hfill
    \subfloat[Color Histogram]{
		\includegraphics[width=0.230\textwidth]{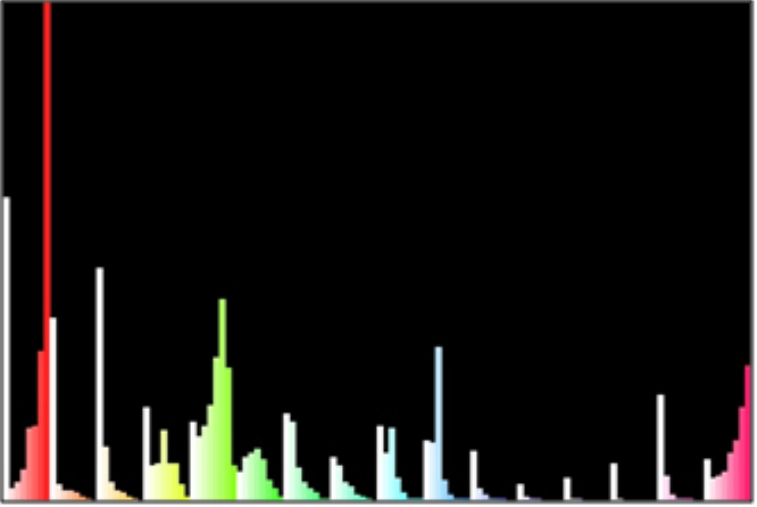}}
	\subfloat[SIFT]{
        \includegraphics[width=0.230\textwidth]{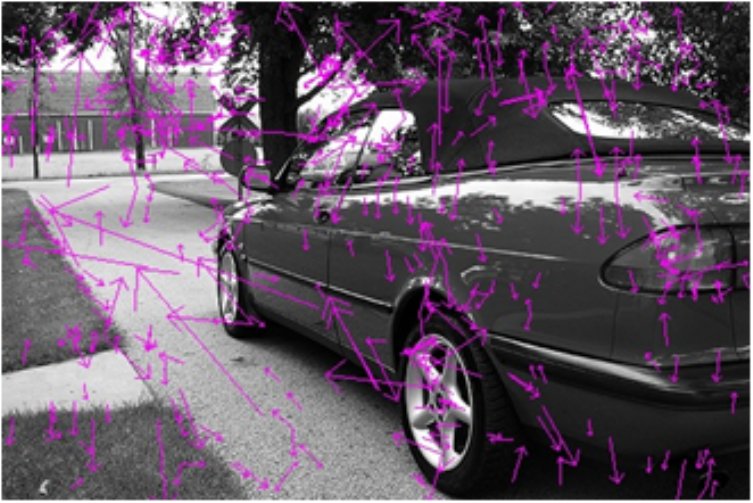}}
    % \vspace{-1em}
    \caption{(a) An image example from VOC 2007. It can be described with different kinds of descriptors, such as: (b) HOG (c) Color histogram and (d) SIFT.}
	\label{fig_car}
	\vspace{-1em}
\end{figure}

High dimensional data pose a lot of challenges for data mining and pattern recognition \cite{wu2014data}. Usually, feature selection is utilized in order to reduce dimensionality by eliminating irrelevant and redundant features \cite{Guyon2003}. In most contexts, feature selection models are oriented to offline situation. That is, the global feature space has to be obtained in advance \cite{liu2007computational}\cite{yu2004efficient}. However, in real-world applications, the features are actually generated dynamically. For %instance
example, in image analysis \cite{wang2012multimodal}, multiple descriptors are exacted to capture various visual information of images, such as Histogram of Oriented Gradients (HOG), Color histogram and Scale-Invariant Feature Transform (SIFT) as shown in Figure \ref{fig_car}. %The visualization of the descriptor of an image is shown in Figure~\ref{fig_car}, each of which is generated independently. 
It is very time-consuming to wait for the calculation of all the features. Thus, it is necessary to perform feature selection by their arrival, which is referred to as online feature selection. The main advantage of online feature selection is its time efficiency and suitable for online applications, therefore, it has emerged as an important topic.

Online feature selection assumes that features flow into the model one by one dynamically. The feature selection is performed by the arrival of features. It is different from classical online learning, in which the feature space remains consistent while samples flow in sequentially \cite{mairal2010online}\cite{bottou1998online}\cite{hoi2012online}\cite{gao2014soml}. There are some papers that focus on this direction \cite{perkins2003grafting} \cite{alpha05}\cite{wu2010online}\cite{wu2013online}. Perkins et al. proposed gradient descent model, Grafting \cite{perkins2003grafting}. It selects features by minimizing the predefined binomial negative log-likelihood loss function. Zhou et al. introduced a streamwise regression model to evaluates the dynamic feature \cite{alpha05}. Wu et al. performed online selection by relevance analysis \cite{wu2013online}. These approaches can evaluate features dynamically with the arrival of each new feature, but they suffer from a common limitation: they overlook the relationship between features which is very important in some real-world applications \cite{shen2014unified}\cite{wang2013view}\cite{wang2012assistive}\cite{zhao2012cooperative}.

% As shown in Figure~{\ref{fig_car}},
In image processing, each kind of cues of the image describes certain information and consists of high dimensional feature spaces. %More Specifically, specifically, the HOG and the SIFT descriptors represent the shape information, the color histogram denotes the color information.
  In bioinformatics, DNA microarray data consist %consists
  of groups of gene sets in terms of biological meanings. The group information can be considered as a type of prior knowledge on the connection of the features, and it is difficult to be discovered from merely data and labels. Therefore, performing selection on feature groups can perform better than perform selection on features individually. Hence, there are some works which focus on feature selection with group structure information, such as group Lasso and sparse group Lasso \cite{groupLasso} \cite{overlapLasso} \cite{zhou2014parallel} \cite{xiang2014simultaneous} \cite{wang2014two}. However, these methods are performed in a batch manner. Although Yang et al. \cite{yang2010online} proposed an online group lasso method, it is designed for instance stream. A global feature space of the data sets is still desired in advance for feature selection.%. It is for the classical online learning which still requires the global feature space of each sample.

Therefore, we first formulate the problem as online group feature selection. There are two challenges for this problem: 1) the features are generated dynamically; 2) they are with group structure. To the best of our knowledge, none of existing feature selection methods can well handle these two issues. Therefore, in this paper, we propose a novel feature selection method for this problem, namely Online Group Feature Selection (OGFS) \cite{wang2013online}. More specifically, on time step $t$, a group of feature $g_t$ is generated. We develop a novel criterion based on spectral analysis which aims to select discriminative features in $g_t$. The process is called an online intra-group selection. Each feature in $g_t$ is evaluated individually in this stage. Then after the intra-group selection on $g_t$ is finished, we reevaluate all the selected features so far to remove the redundancy. The process can be accomplished with a sparse linear regression model, Lasso. We refer this stage as an online inter-group selection. Our major contributions are summarized as follows:
\begin{itemize}
\item To the best of our knowledge, this is the first effort that considers the group structure in an online fashion. Although online feature selection methods are proposed, we here utilize the group structure information in the feature stream.

\item  Based on the observation that spectral analysis is widely used for discriminative variable analysis, we propose a novel criterion based on spectral analysis. The criterion is proven to be efficient in the online intra-group feature selection.

\item To get benefit from the correlation among features from groups, we use a sparse regression model Lasso for the online inter-group feature selection. It is the first time that the sparse model Lasso is employed in the dynamic feature selection.%Our proposed Online Group Feature Selection (OGFS) method achieves the best performance conducted on the real-world and benchmark data sets compared to the state-of-the-art online feature selection methods. That demonstrate that our method is more suitable for real-world applications.%with other state-of-the-art methods for online feature selection on real-world and benchmark data sets.
\item
We demonstrate the superiority of our method over the state-of-the-art online feature selection methods. The experimental results on real-world applications show the effectiveness of our method for tasks with large scale data, such as image classification and face analysis.
\end{itemize}
The online group feature selection was first introduced in our previous work~\cite{wang2013online}. In comparison with the preliminary version \cite{wang2013online}, we have improvements in the following aspects: (1) we performed a more comprehensive survey of existing related works; (2) to solve regression sparse model in inter-group selection, we adopted a more efficient solution; (3) we conducted more empirical evaluations; and (4) more discussions and analysis are provided.
The rest of the paper is organized as follows. After review of related work in Section \ref{sec2}, we introduce our framework and give our algorithm in Section \ref{framework}. Then we report the empirical study on real-world and benchmark 
data sets in Section \ref{experim}. Section \ref{conclu} concludes this paper and discusses possible future work.

\section{Related Work}
\label{sec2}
In this section, we first give a brief review of traditional offline feature selection, including filter, wrapper and embedded models. Specifically, we review existing literature that focus on utilizing the underlying group structure of feature space, such as group Lasso and their extensions. Then, we introduce the state-of-the-art online feature selection methods.%Then, we introduce the state-of-the-art online feature selection methods.

\subsection{Offline Feature Selection}
Traditional feature selection is oriented to the off-line situation. The problem statement is defined below. Given a data set %sets
$X=[x_1, x_2, \cdots, x_n]\in R^{n\times d}$ consisting of $n$ samples (columns) over $d$-dimensional feature space $F=[f_1, f_2, \cdots, f_d]\in R^d$, pre-process of the features such that each row is centered around zero and is of unit $L_2$ norm $||f_i||=1$. The object of feature selection is to choose a subset of features $S\in R^l$ from the global feature space $F$, and $l$ is the desired number of features, and in general $l < d$.

Generally, the feature selection methods fall into three classes based on how the label information is used. Most existing methods are supervised which evaluate the correlation among features and the label variable. Due to the difficulty in obtaining labeled data, unsupervised feature selection has attracted increasing attention in recent years \cite{dy2004feature}. Unsupervised feature selection methods usually select features that preserve the data similarity or manifold structure \cite{yang2011}. Semi-supervised feature selection, so called ``small-labeled sample problem'', makes use of label information and manifold structure corresponding to labeled data and unlabeled data \cite{zhao2007spectral}.

The existing feature selection methods can be categorized as embedded, filter and wrapper approaches based on the methodologies \cite{farahat2012efficient}\cite{zhao2013similarity}\cite{filter2001}\cite{wu2010online}\cite{wang2007trace}. The filter methods evaluate the features by certain criterion and select features by ranking their evaluation values. The correlation criteria proposed for feature selection include mutual information, maximum margin \cite{guyon2003introduction}, kernel alignment \cite{combarro2005introducing}, and the Hilbert Schmidt independence criterion \cite{song2007supervised}. The development of filtering methods involves taking consideration of multiple criteria to overcome redundancy. The most representative algorithm is mRMR \cite{peng2005feature} in the principle of max-dependency, max-relevance and min-redundancy. %More specifically, mRMR
It aims to find a subset in which the features are with large %larges
dependency on the target class and with low redundancy among each other.

The wrapper methods employ a specific classifier to evaluate a subset directly. For example, Weston et al. \cite{guyon2002gene} used SVM as a wrapper with the purpose of optimizing the SVM accuracy on each subset of features. %Though
The wrapper methods usually have better performance than the filter methods. However, they are typically computationally expensive, as
the time complexity is exponential with respect to the number of features. Meanwhile, %In the same time,
the performance of the selected subset relies on the specific training classifier. %certain training classifier.

The embedded methods usually seek the subset by jointly minimizing empirical error and penalty. They tend to be more efficient than the wrapper % wrapper
model and have a relatively small % small
size of the ultimate subset. %, such as the Lasso %quadratic 1-norm SVM \cite{weston2000feature}.
 LARS is a successful example that falls into this category \cite{efron2004least}. Its objective function is to minimize %reconstruction
 the reconstruction error with sparsity constraint on the coefficients of the features. The sparsity constraint can lead to a small number of nonzero estimates. There are also some generalized methods, such as adaptive Lasso \cite{zou2006adaptive} and group Lasso \cite{yuan2011efficient}. % greedy Lasso %\cite{zhang2009consistency} and so on.

 We take group Lasso as an example. It considers the correlation structure in the feature space. The underlying structure in feature space is important in feature selection. Take the application of bioinformatics as an example, certain factors which contribute to predicting the cancer consist of a group of variables. Then, the problem amounts to the selection of groups of variables. Group Lasso and its extended works mainly solve the following optimization problem:
 \begin{equation}
 \label{glasso}
% \begin{split}
  \min_{\w} L(\w)
+\lambda_1||\w||_1+\lambda_2 \sum_{i=1}^{G}\beta_i||\w_{G_i}||_2 ,
%  \end{split}
 \end{equation}
where $L(\cdot)$ is a smooth convex loss function such as the least squares loss. The feature space is partitioned into $G$ groups ${x_{G_i}}$, $\beta >0$ is the parameter corresponding to each group, and $\lambda_1\geq 0$ and $\lambda_2\geq 0$ are regularization parameters which modulate %control
the sparsity of the selected features and groups respectively.
When parameters $\lambda_1$ and $\lambda_2$ are set to different values, the model (\ref{glasso}) falls into the different models as seen in Table~\ref{vlasso}. %missing £¿£¿£¿For overlapping groups, there are

\begin{table}[t]
\caption{General group Lasso model with various parameters}
 \vspace{-0.2in}
\label{vlasso}
\begin{center}
\begin{tabular}{cll} %llll
\hline
\multicolumn{1}{l}{\bf Parameters }
&\multicolumn{1}{l}{\bf Group}
&\multicolumn{1}{l}{\bf Algorithm}
\\ \hline
$\lambda_1>0,~\lambda_2=0$         &Unique &Lasso \cite{lasso}\\ \hline
$\lambda_1=0,~\lambda_2>0$         &Disjoint &group Lasso \cite{groupLasso}\\ \hline
$\lambda_1>0,~\lambda_2>0$             &Disjoint &sparse group Lasso \cite{overlapLasso}\\
             &Overlapping &overlapping group Lasso \cite{yuan2011efficient} \\ \hline
\end{tabular}
 \vspace{-0.1in}
\end{center}
\end{table}
Yang et al. \cite{yang2010online} proposed an online algorithm for the group Lasso. The weight vector $\w$ is updated by the arrival of a new sample. Important features corresponding to large values in $w$ are selected in a group manner. Thus, the algorithm is suitable for sequential samples, especially for the applications with large scale data.

The aforementioned feature selection methods are offline or designed for the classical online scenario, in which instances arrive dynamically instead of features. There are some works focus in this aspect. A brief review is summarized in the next subsection.
%There are some research focusing on the feature selection of the feature stream. A brief review is summarized in the following section.

\subsection{Online Feature Selection}
%In this paper,
Online feature selection assumes that features arrive in by streams. It is different from classical online learning which lets samples flow in dynamically. Thus, at time step $i$, there is only one feature descriptor $f_i$ of all samples available. %mini-batch of features $F=[f_1, \cdots, f_i]\in R^{n\times b}$ of all samples available, where $n$ is the number of samples and $b$ is the batch size.
The goal of online feature selection is to justify whether the feature $f_i$ should be accepted by their arrival. To this end, some related works have been proposed, including Grafting \cite{perkins2003grafting}, Alpha-investing \cite{alpha05} and OSFS (Online Streaming Feature Selection) \cite{wu2013online}. %Details are below. %, %OGFS (Online Group Feature Selection) \cite{wang2013online} and so on.
\subsubsection{Grafting}
Grafting integrates the feature selection in learning a predictor within a regularized framework. Grafting is oriented to binomial classification, its objective function is a binomial negative log-likelihood loss (BNLL) function, defined as:
\begin{equation}
\label{graft}
\min_{W} \frac{1}{n}\sum_{i=1}^{n}ln(1+e^{-y_if(x_i)}) + \lambda \sum_{j=1}^{k}|w_j|_1,
\end{equation}
 where $n$ is the number of samples and $k$ is the number of selected features so far, the predictor $W$ is constrained with the $L_1$ regularization. Note that if a feature $f_j$ is included, $\lambda \|w_j\|$ is penalized. To guarantee the decrease of objective function, the reduction in the mean loss of $\overline{L}$ should outweigh the regularizer penalty to $\lambda \|w_j\|$. Therefore, to justify whether the inclusion of the feature can improve the existing model, Grafting uses a gradient-based heuristic. The feature $f_j$ can be selected if the following condition is satisfied:
\begin{equation}
\label{grad}
\mid \frac{\partial \overline{L}}{\partial w_j} \mid > \lambda,
\end{equation}
where $\lambda$ is a regularization coefficient. Otherwise the weight is dropped and the feature is rejected. Each time a new feature is selected, the model goes back and reapplies the gradient test to features selected so far. %time steps.
The framework is adaptive for both linear and non-linear models. %Though
Grafting has been successfully employed in some applications, such as edge detection \cite{glocer2005online}.
But there are some limitations below. First, though Grafting can obtain a global optimum with respect to features included in the model, it is not optimal as some features are dropped during online selection. Besides, the gradient retesting over all the selected features greatly increases the total time cost. %the processing time. The last
Last, tuning a good value for the important regularization parameter $\lambda$ requires the information of the global feature space.

\subsubsection{Alpha-investing}
Alpha-investing \cite{alpha05} belongs to the penalized likelihood ratio methods \cite{pillers2002mathematical} which do not require the global model. More specifically, for feature $f_i$ arriving at time step $i$, Alpha-investing evaluates it by the p-statistic which leads to $p$-value. The $p$-value is the probability that the feature could be accepted while it should be actually discarded. Then comparing the $p$-value of $f_i$ with the threshold $\alpha_i$, the feature $f_i$ is added to the model if its $p$-value is greater than $\alpha_i$. The threshold $\alpha_i$ corresponds to the probability of including a spurious feature at time step $i$. Each time a feature is added, the wealth $w_i$ will increase as shown in Eq.~\ref{wi+1}, where $w_i$ represents the current acceptable number of future false positives.
\begin{equation}
\label{wi+1}
w_{i+1} = w_i+\alpha \Delta - \alpha_j.
\end{equation}
Otherwise, the feature $f_i$ is discarded and $w_i$ will decrease as shown in Eq.~\ref{wi+1r}.
\begin{equation}
\label{wi+1r}
w_{i+1} = w_i - \alpha_i,
\end{equation}
where $\alpha \Delta$ is the parameter controlling the false discovery rate, and $\alpha_i$ is set to be $w_i/(2\times i)$ at time step $i$. In summary, Alpha-investing adaptively adjusts the threshold for feature selection. It can also handle an infinite feature stream. However, Alpha-investing does not reevaluate the included features which will greatly influence the following selection.

\subsubsection{OSFS}
In OSFS, %The main strategy of is based on the\cite{john1994irrelevant},
features are characterized as strongly relevant, weakly relevant, or irrelevant\cite{john1994irrelevant} with the label attribute. With the incoming %coming%
of new feature $f_i$ at time step $i$, OSFS first analyzes its correlation with the label $y$. If the feature is weakly or strongly relevant to the label, it will be selected. %Meanwhile, to reduce the redundancy among existing features, an online Markov blanket is further developed. %Then, Wu et al. developed an online Markov blanket to reduce the redundancy among existing features.
%More precisely, %That is,
%they built
%the Markov blanket of class label MB($Y$) is built over time. Each time a feature $f_j$ is selected, online Markov blanket determines
 If the feature $f_i$ is added, OSFS performs redundancy analysis. That is, in the condition of selecting a new feature, some previously selected features become irrelevant and will also be removed. More specifically, a feature $f_i$ is redundant to the class feature $y$ if it is weakly relevant to $y$. Let MB($f_i$) denote a Markov blanket of $y$, which is a subset of MB($y$) containing all the weakly or strongly relevant but non-redundant features. Thus, redundancy analysis is a key component for an optimal feature selection process. OSFS does not need parameter tuning and shows outstanding performance in many applications, such as impact crater detection.

All the above methods are state-of-the-art online feature selection methods. Although existing methods greatly relieve the burden of processing high dimensional data sets, they do not consider the correlation among features. Hence, we address the online group feature selection problem in this work. To make use of the prior knowledge of group information, we propose an efficient online feature selection framework including the intra-group feature selection and inter-group feature selection. Based on this framework, we develop a novel algorithm called Online Group Feature Selection (OGFS).

\section{Online Group Feature Selection}
\label{framework}
We first formalize our problem for online group feature selection.
Assume a data matrix $X=[x_1,\cdots,x_n]\in R^{d \times n}$, where $d$ is the number of features arrived so far and $n$ is the number of data points, and a class label vector $Y=[y_1,\cdots,y_n]^{T}\in \mathbb{R}^{n}$, $y_i\in\{1,\cdots,c\}$, where $c$ is the number of classes. The feature space is a dynamic stream vector $F$ consisting of groups of features, $F=[G_1,\cdots,G_j,\cdots]^{T}\in \mathbb{R}^{\sum d_{j}}$, where $d_{j}$ is the number of features in group $G_{j}$. $G_j=[f_{j1},f_{j2},\cdots,f_{jd_j}]^{T}\in R^{d_j}$ where $f_{jk}$ is an individual feature.
In terms of feature stream $F$ and class label vector $Y$, we aim to select an optimal feature subset $U=[g_1,\cdots,g_j,\cdots, g_k]^{T}\in \mathbb{R}^{\sum k_{j}}$ when the algorithm terminates, where $g_j\in \mathbb{R}^{k_j}$ is the selected feature space from $G_j$, that is, $g_j\subseteq G_j$. $k_{j}$ is feature dimension of $g_j$, $0\leq k_j\leq d_j$.
\begin{figure}
\centering
 \includegraphics[width=0.45\textwidth]{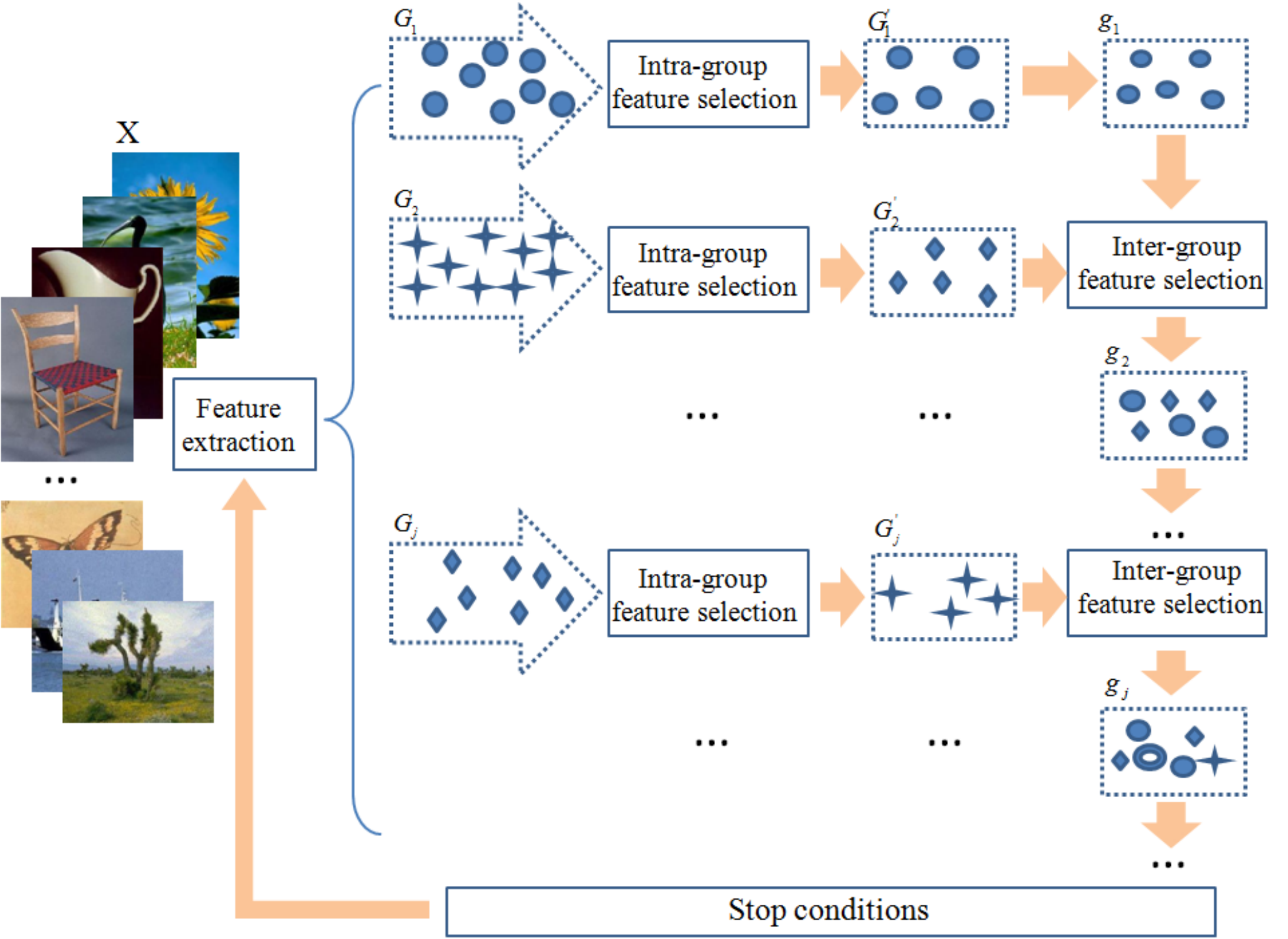}
 \caption{Schematic illustration of the online group feature selection approach.}%The procedure of OGFS in online scenario.}
 \label{diagram}
 \vspace{-0.1in}
\end{figure}

To solve this problem, we propose a framework for online group feature selection which consists of two components: intra-group selection and inter-group selection. The intra-group selection is to process each feature dynamically at its arrival. That is, when a group of features $G_j$ is generated, we process the feature individually and select a subset $G_j'$. In terms of the features obtained by the intra-group selection $G_j'$, we further consider the correlation among the groups and get an optimal subset $g_j$, namely the inter-group selection. The overview of the procedure is illustrated in Figure~\ref{diagram}. Based on this framework, we propose a novel Online Group Feature Selection (OGFS). In the following subsections, we will give details of our algorithm.

 \subsection{Online Intra-Group Selection}

Spectral based feature selection methods have demonstrated their effectiveness \cite{redundancy}. Given a data matrix $X\in \mathbb{R}^{d\times n}$, we construct two weighted undirected graphs $G_w$ and $G_b$ on given data. Graph $G_w$ reflects the within-class or local affinity relationship, and $G_b$ reflects the between-class or global affinity relationship. The Graphs $G_w$ and $G_b$ are characterized by the weight matrices $S_w$ and $S_b$, respectively. The weight matrices $S_w$ and $S_b$ can be constructed to represent the relationships among instances, such as RBF kernel function. % we construct a weighted graph with edges between data points close to each other.is constructed. Let $S_b\in \mathbb{R}^{n\times n}$ evaluate the between-class distance, and $S_w\in \mathbb{R}^{n\times n}$ evaluate the within-class distances.
In this work, we only consider supervised online feature selection. The between-class adjacency matrix $S_b$ and the within-class adjacency matrix $S_b$ are calculated as follows \cite{trace2008}:
\begin{equation}\label{sb}
(S_{b})_{ij}=
\begin{cases}
\frac{1}{n}-\frac{1}{n_l},\quad& y_i=y_j=l, \\
\frac{1}{n}, \quad	 \quad & y_i\neq y_j.
\end{cases}
\end{equation}
\begin{equation}\label{sw}
(S_{w})_{ij}=
\begin{cases}
\frac{1}{n_l},\quad& y_i=y_j=l, \\
0, \quad	 \quad & y_i\neq y_j.
\end{cases}
\end{equation}
where $n_l$ denotes the number of data points from class $l\in {1,\cdots,c}$. Given the adjacency matrix $S_w$ and $S_b$, we introduce the definitions of degree matrix and Laplacian matrix which are frequently used in spectral graph theory.

\newtheorem{meth}{Definition }

\begin{meth}
(Degree matrix) Given the adjacency matrix $S_w$ of the graph $G_w$, the degree matrix $D_w$ is defined by: $D_{w}=\text{diag}(S_w\bm{1})$ if $i=j$, and 0 otherwise. Similarly, given the adjacency matrix $S_b$ of the graph $G_b$, the degree matrix $D_b$ is defined by: $D_{b}=\text{diag}(S_b\bm{1})$ if $i=j$, and 0 otherwise. $\bm{1}$ is an identity vector.
\end{meth}

According to the definition, the degree matrix is a diagonal matrix. $D_w(ii)$ can be interpreted as an estimation of the density around the node $x_i$ in graph $G_w$, same as $D_b(ii)$.

\begin{meth}
(Laplacian matrix) Given the adjacency matrix $S_w$ and the degree matrix $D_w$ of the graph $G_w$, the Laplacian matrix of graph $G_w$ is defined as $L_w=D_w-S_w$. Similarly, the Laplacian matrix of graph $G_b$ is defined as $L_b=D_b-S_b$.
\end{meth}

The degree matrix and the Laplacian matrix satisfy the following property \cite{chung1997spectral}: $\forall x\in R^d$, $x^TL_wx=\sum_{ij}|||x_i-x_j||^2(S_w)_{ij}$, similarly $x^TL_bx=\sum_{ij}|||x_i-x_j||^2(S_b)_{ij}$.

Applying the spectral graph theory to feature selection, it is about finding a smooth feature selector matrix which is consistent with the graph structure. Let $W=[w_i,\cdots,w_m]^{T}\in \mathbb{R}^{d\times m}$ denote the feature selector matrix, where $d$ is the number of features selected and $m$ is the dimension of global feature space. Here $w_i=[w_{i1},\cdots,w_{id}]^{T}\in \mathbb{R}^{d}$ has only one entry $w_{ii}$ equal to ``1''. With the procedure of feature selection, the data matrix $X$ is transformed to $Z\in R^{m\times n}$ by the feature space projection, $Z=W^TX$.

In the feature space indicated by a smooth selection matrix $W$, the instances of the same class are close to each other on $G_w$. In the same time, the instances of different classes are distant from each other on $G_b$. $S_w$ reflects the the within-class or local affinity relationship. Specifically, if data $x_i$ and $x_j$ belong to the same class or are close to each other, $S_w(ij)$ is a relatively larger value. Otherwise $S_w(ij)$ is a relatively smaller value. Therefore, we should select the feature subset that makes $\sum_{ij}||z_i-z_j||^2S_w(ij)$ as small as possible. Similarly, $S_b$ reflects the between-class or global affinity relationship. If instances $x_i$ and $x_j$ belong to the different classes, $S_b(ij)$ is a relatively larger value. Therefore, we should select the feature subset which ensures that $\sum_{ij}||z_i-z_j||^2S_b(ij)$ is as large as possible. To sum up, the best selection matrix can be achieved by maximizing the following objective function:
\begin{equation}\label{max_pre}
      F(W_U)=\frac{\sum_{ij}||z_i-z_j||^2S_b(ij)}{\sum_{ij}||z_i-z_j||^2S_w(ij)}.
\end{equation}
With the property of Laplacian matrix, we obtain the following equivalent program:
\begin{equation}
 \begin{split}
     \sum_{ij}||z_i-z_j||^2S_b(ij) &= {Z}L_b{Z}^T \\
     &= W_U^TXL_bXW_U.
 \end{split}
\end{equation}
Similarly, we can get $\sum_{ij}||z_i-z_j||^2S_w(ij) = W_U^TXL_wXW_U$. The objective function of \ref{max_pre} can be transformed as: % Spectral feature selection approaches can be categorized into subset-level selection and feature-level selection approaches. More specifically, the %The subset-level selection is to find an optimal subset $U$ by maximizing the following criterion:
\begin{equation} \label{max}
      F(U)=\frac{\text{tr}(W_U^T(XL_bX^T)W_U)}{\text{tr}(W_U^T(XL_wX^T)W_U)},
\end{equation}
%where $W_U$ corresponds to the features in subset $U$, $L_{b}$ and $L_{w}$ are the Laplacian matrices, %\cite{chung1997spectral}$L_{b}=D_{b}-S_{b}$, $D_{b}$ is a diagonal matrix, $D_{b}=\text{diag}(S_{b}\bm{1})$; $L_{w}=D_{w}-S_{w}$, $D_{w}$ is a diagonal matrix and $D_{w}=\text{diag}(S_{w}\bm{1})$.

The feature-level spectral feature selection approach evaluates feature $f_i$ by a score defined below:
\begin{equation}\label{fs}
      s(f_i)= \frac{w_i^T(XL_bX^T)w_i}{w_i^T(XL_wX^T)w_i}.
\end{equation}

After obtaining all feature scores, the feature-level approach will select the leading features corresponding to the top ranking scores. %by the rankings of scores.
As traditional spectral feature selection approaches rely on the global information, they are not efficient for online fashion.

Hence, to get benefit from spectral analysis, we evaluate the new arrival feature by the criterion defined by Eq. (\ref{max}). In the Eq. (\ref{max}) of streaming feature scenario, $W^{d \times m}$ denotes the online feature selector matrix, where $d$ denotes the arrived features so far and $m$ denotes the selected features. Given the selected feature space $U$, the new arrival feature $f_i$ will be selected if its inclusion improves the discriminative ability of the feature space, that is:
\begin{equation}\label{crite}
\begin{split}
% \color{blue}
    F(U\cup f_i)-F(U) &> \varepsilon,
%    \text{s.t.} \  &\varepsilon<0.001\\
\end{split}
\end{equation}
where $\varepsilon >0$ is a small positive parameter. However, the performance is easily influenced by the sequence of arriving features. Specifically, if the previous arrived features are with high level of discriminative capacity, it is difficult for the following features to satisfy (\ref{crite}). Thus, we allow the discriminative ability of the feature disturb within the range of $\varepsilon$. Then, the criterion based on spectral analysis for streaming feature scenario is defined as follows.

\newtheorem{met}{Criterion }
\begin{meth}
Given $U\in \mathbb{R}^{b}$ as the previously selected subset, $f_i$ the newly arrived feature, we assume that with the inclusion of a ``good'' feature, the between-class distances will be larger, while the within-class distance will be smaller. That is, feature $f_i$ will be selected if the following criterion is satisfied:
\end{meth}
\vspace{-0.25in}
\begin{equation}\label{dis}
\begin{split}
    |F(U\cup f_i)-F(U)| &> \varepsilon\\
\end{split}
\end{equation}
where we use $\varepsilon=0.001$ in our experiments.

%However, with the increase of selected features, the criterion defined in Eq. (\ref{max}) will be more %and more
%difficult to meet. Hence, to avoid leaving out discriminative features, we design the second criterion.

%\begin{meth}
%Given $U\in \mathbb{R}^{b}$ as the previously selected subset, and the newly arrived feature $f_i$, we calculate the score of feature $f_i$ by Eq. (\ref{fs}) which shows the discriminative power of the feature. If it is a significant feature with discriminative power, it will be selected.
%\end{meth}
%
%The significance of a feature can be evaluated by the $t$-test \cite{ttest} defined bellow:
%\begin{equation}\label{test}
%      t(f_i,U)= \frac{\hat{\mu}-s(f_i)}{\hat{\sigma}/\sqrt{|U|}}
%\end{equation}
%where $|U|$ stands for the number of features in $U$, and $\hat{\mu}$ and $\hat{\sigma}$ are the mean and standard deviation of scores of all the features in $U$ respectively. %.
%If the $t$-value returned by Eq. (6) reaches 0.05, then the feature is considered as %assumed to be significant
%a significant feature among the selected subset $U$ and will be selected ($0.05$ is an empirical threshold used to measure the significance level).

After intra-group selection, we will obtain a subset $G_j'\in \mathbb{R}^{m'}$ from the original feature space $G_j$, $G_j'\subset G_j$. However, the criterion 1 will include discriminative features but may also cause redundancy. Meanwhile, the intra-group selection evaluates the streaming features individually and does not consider the group information. Thus, we further apply inter-group selection. Our inter-group selection is based on the classical sparse model Lasso which could reduce the redundancy among selected features efficiently.
%As intra-group selection evaluates the features individually and does not consider the group information, we will apply inter-group selection.

 \subsection{Online Inter-Group Selection}

In this section, we introduce the online inter-group selection which aims to obtain an optimal subset based on global group information. We propose to solve the problem with a linear regression model, Lasso. Given the subset selected at %during
the first phase $G_j'=[f_{j1},f_{j2},\cdots,f_{jm'}]^{T}\in \mathbb{R}^{m'}$, the previously selected subset of features $U^{T}\in \mathbb{R}^{b}$, the combined feature space with dimension of $m''$ ($m'+d = m''$), a data set matrix $X \in \mathbb{R}^{m''\times n}$, and a class label vector $y\in \mathbb{R}^{n}$, $\hat{\beta}=[\hat{\beta}_1,\cdots,\hat{\beta}_{m''}]\in \mathbb{R}^{m''}$ is the projection vector which constructs the predictive variable $ \hat{y}$:
 \begin{equation}\label{linear}
  \hat{y} = X^{T}\hat{\beta}.    \\
\end{equation}
the sparse regression model Lasso chooses an optimal $\hat{\beta}$ by minimizing the objective function defined as follows:
\begin{equation}\label{lasso}
    \begin{split}
   & \min_{\hat{\beta}} \  ||y-\hat{y}||_2 ,  \\
    \text{s.t.} & \   ||\beta||_1\leq\lambda,~\hat{y} =X^{T}\hat{\beta},
      \end{split}
\end{equation}
where $||\cdot||_2$ stands for $l_2$ norm, and $||\cdot||_1$ stands for $l_1$ norm of a vector, $\lambda$ is a parameter that controls the amount of regularization applied to estimators, and $\lambda\geq0$. In general, a smaller $\lambda$ will lead to a sparser model. To solve the problem defined in Eq.~(\ref{lasso}), we reformulate the function as:
\begin{equation}\label{nolasso}
 \min_{\hat{\beta}} \  ||y-X^{T}\hat{\beta}||_2+ \lambda||\hat{\beta}||_1,
\end{equation}
which can be solved efficiently by many optimization methods such as feature-sign search \cite{lee2007efficient}. In the optimization methods, the value of $\lambda$ is usually determined by cross validation. % First, find the features which locally improve %improves
%the objective function into an %a
%active feature subset and their corresponding signs. Second, %Then
%search for the optimal active set. The procedure will continue until the optimization function is satisfied.
The sparse regression model selects features by setting several component in $\beta_i$ to zero, then the corresponding feature $f_i$ is deemed to be irrelevant to the class label and should be discarded. Finally, the features corresponding to non-zero coefficients will be selected.

After inter-group selection, we get the subset $U_j$.
%optimization function are satisfied.
 With the combination of the online intra-group and the inter-group selection, the algorithm of Online Group Feature Selection (OGFS for short) can be formed. %OGFS can benefit from intra-group selection in selecting discriminative features and inter-group selection in compactness control.

 \subsection{OGFS: Online Group Feature Selection Algorithm}
\label{ogfs33}
Algorithm 1 shows the pseudo-code of our online group feature selection (OGFS) algorithm. OGFS is divided into two parts: intra-group selection (Step 4-15) and inter-group selection (Step 16). Details are as follows.

In the intra-group selection, for each feature $f_i$ in group $G_j$, we evaluate features by the criterion defined in Section 3.1. Steps (9-11) evaluate the significance of features based on Criterion 1. With the inclusion of the new feature $f_i$, if the within-class distance is minimized and the between-class distance is maximized, feature $f_i$ is considered to be a ``good'' feature and will be added to $G_j'$. If the inclusion of the new feature $f_i$ causes the discriminative ability of the feature space disturb in a arrange $\varepsilon$, it may be helpful and also selected.%Steps (12-14) evaluate the features according to Criterion 2.
%Based on the selected subset $U$, we validate the significance of the feature by $t$-test. If the $t$-value returned by Eq. (6) is larger than 0.05, feature $f_i$ is thought to be significant in discrimination. Then $f_i$ will be added to $G_j'$.
After intra-group selection, we get a subset of features $G_j'$. To implement the global information of groups, we build a sparse regression model based on the selected subset $U$ and the newly selected subset $G_j'$. An optimal subset $g_j$ will be returned by the objective function defined in formula~\ref{lasso}.%Eq. (\ref{lasso}).

In our algorithm, the selected features will be re-evaluated %reevaluated
in the intra-group selection in %in
each iteration. The time complexity of intra-group selection is $O(m)$, and the time complexity of inter-group selection is $O(q)$. Therefore, the time complexity of OGFS is linear with respect to the number of features and the number of groups. %whose time complexity is linear with the number of features and the number of groups, is very fast.

The above iterations %iterations
will continue until the performance of $\psi(U)$ reaches a predefined threshold below.%as follows:
 \begin{itemize}
 \item $|U|\geq k$, $k$ is the number of features we need to select;
 \item $accu(U)\geq max$, the prediction %predictive
 accuracy of the model based on $U$ reaches the predefined accuracy $max$;
 \item There are no more yet-to-be-coming features. % to come.
\end{itemize}
{
\renewcommand\algorithmicrequire{\textbf{Input:}}
\renewcommand\algorithmicensure {\textbf{Output:}}
\begin{algorithm}[!t]
\caption{OGFS (Online Group Feature Selection)}
\begin{algorithmic}[1] %\begin{algorithmic}[1]Êý×Ö1±íÊ¾Ã¿Ò»ÐÐÒ»¸öÐòºÅ
\REQUIRE  feature stream $F\in \mathbb{R}^{m \times q}$, label vector $Y\in \mathbb{R}^{n}$.
\ENSURE selected subset $U$.
\STATE{ $U=$[], $i=1$, $j=1$;}
\STATE {\textbf{while} $\psi(U)$ not satisfied \textbf{do}}
\FOR{$j=1$ to $q$}
\STATE $G_j \leftarrow$ generate a new group of features;
\FOR{$i=1$ to $m$}
\STATE $G_j' =$ [];
\STATE $f_i\leftarrow$ new feature;
\STATE /***evaluate feature $f_i$ by criterion 1, 2***/
\IF  {$ F(f_i\bigcup G_j')- F(G_j')>\varepsilon $}
\STATE  $G_j' = G_j'\bigcup f_i$;
%\STATE  where $F(f_i\bigcup G_j^{'})$, $F(G_j^{'})$ are defined in Eq. (\ref{max});
\ENDIF
%\IF  {$ t(f_i, U) > 0.05 $}
%\STATE  $G_j' = G_j'\bigcup f_i$;
%\STATE  where $t(f_i,U)$ is defined in Eq. (6);
%\ENDIF
\ENDFOR
\STATE {$g_j\leftarrow$ find the global optimal subset $G_j'$ by the feature-sign search algorithm};
\STATE {$U = U \bigcup g_j$}
\ENDFOR
\STATE {\textbf{end while}}
\end{algorithmic}
%\textbf{end while}
\end{algorithm}
}

\section{Experiments}
\label{experim}
In this section, we empirically show the superiority of our method. In experimental settings, we present the comparative methods, evaluation metrics and the simulation of online situation. Then encouraging results on real-world applications such as image classification and face verification are reported. We will verify the influence of group orders in our OGFS method. We also conduct experiments on UCI benchmark data sets to further verify the effectiveness of our method.

\begin{figure*}
\centering
   \subfloat[Cifar-10]{
        \includegraphics[width=0.33\textwidth]{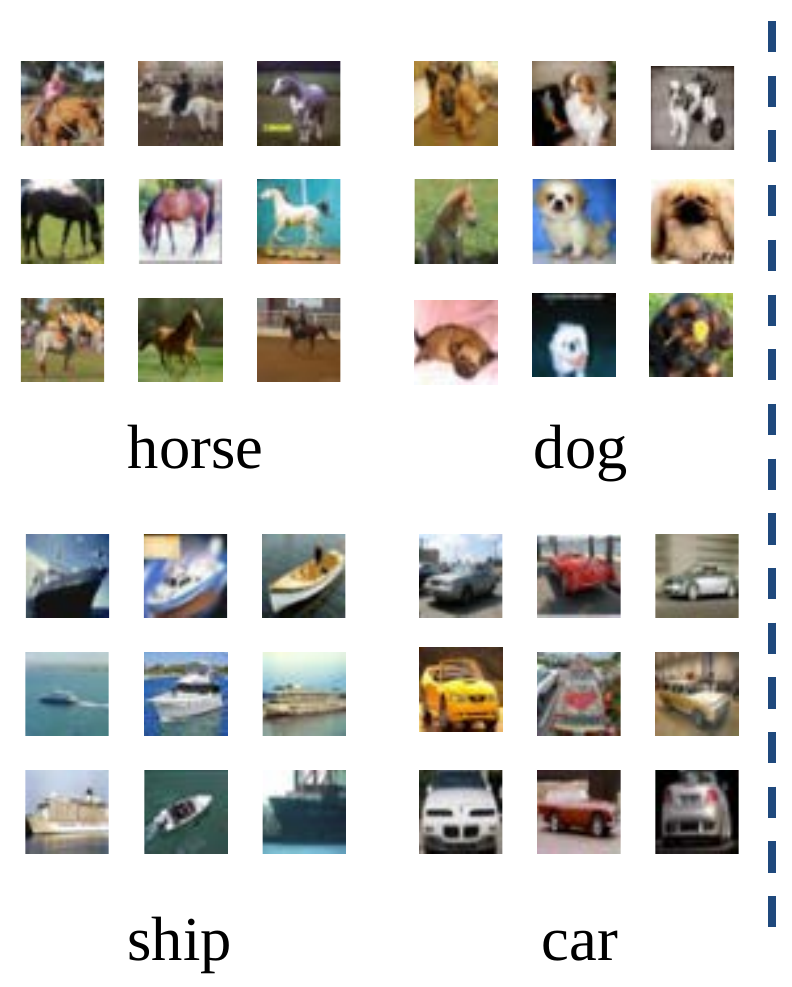}}
     \subfloat[Caltech-101]{
        \includegraphics[width=0.63\textwidth]{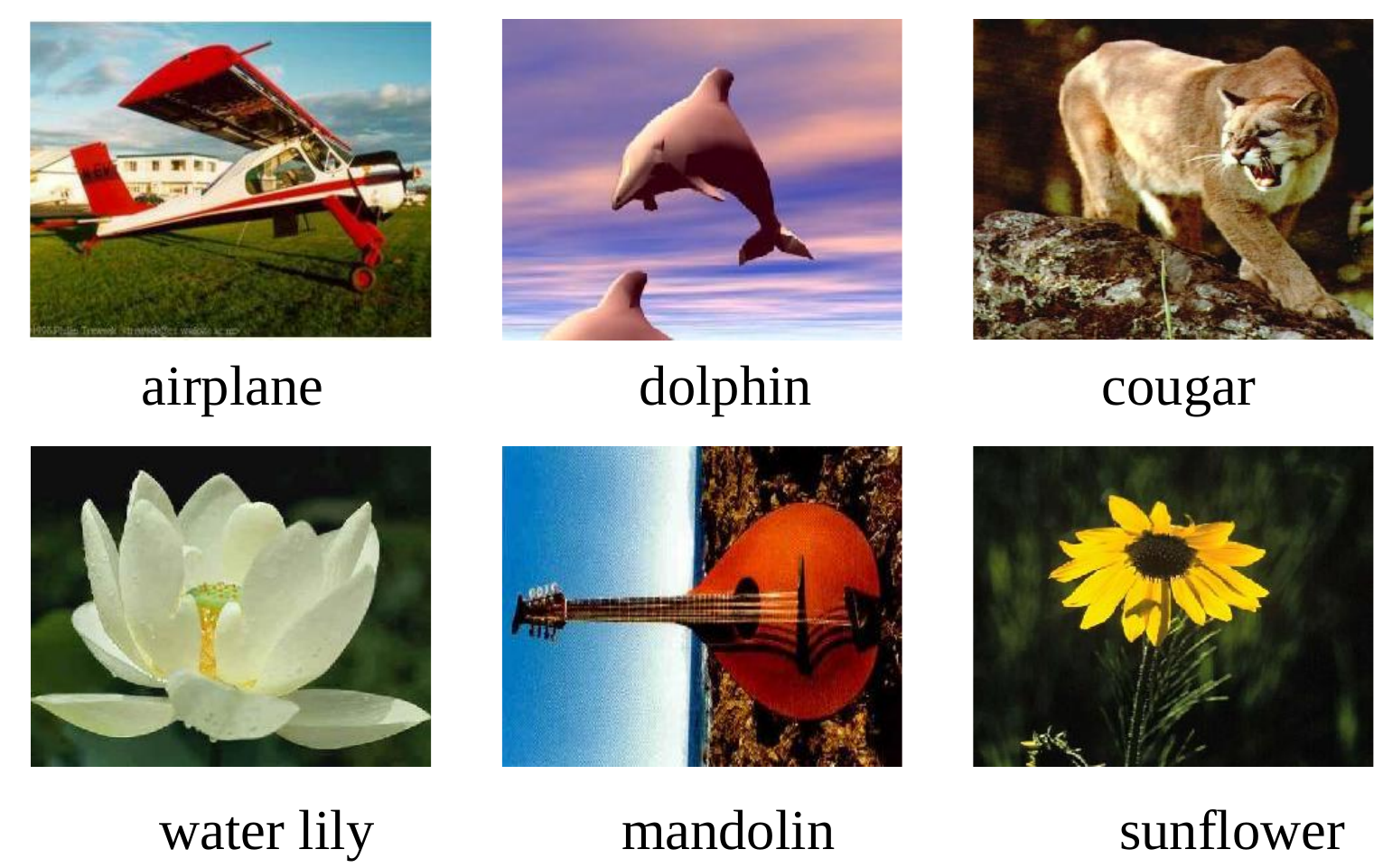}}
%    % \vspace{-1em}
    \caption{Example images from the (a) Cifar-10 and (b) Caltech-101 data sets.}
    %       \label{obj_rec}
	\vspace{-1em}
\end{figure*}
%\begin{figure*}
% \includegraphics[width=1.0\textwidth]{obj.pdf}
%   \caption{Example images from the (a) Cifar-10 and (b) Caltech-101 data sets.}
%          \label{obj_rec}
%\end{figure*}
%We divide this section into three subsections, including an introduction to our experimental settings, the experimental comparison conducted on real-world and the benchmark data sets. Details are as follows.[horse]
%\begin{figure*}
%   \subfloat {
%       \includegraphics[width=0.2\textwidth]{horse.pdf}}
%   \subfloat[dog]{
%        \includegraphics[width=0.20\textwidth]{dog.pdf}}
%   \subfloat[airplane]{
%        \includegraphics[width=0.20\textwidth]{b1.pdf}}
%   \subfloat[dolphin]{
%        \includegraphics[width=0.20\textwidth]{b2.pdf}}
%   \subfloat[cougar]{
%        \includegraphics[width=0.20\textwidth]{b3.pdf}}\\
%%        \hfill
%    \subfloat[ship]{
%		\includegraphics[width=0.2\textwidth]{ship.pdf}}
%	\subfloat[automobile]{
%        \includegraphics[width=0.2\textwidth]{car2.pdf}}
%    \subfloat[water lily]{
%        \includegraphics[width=0.2\textwidth]{b4.pdf}}
%    \subfloat[mandolin]{
%        \includegraphics[width=0.2\textwidth]{b5.pdf}}
%    \subfloat[sunflower]{
%        \includegraphics[width=0.2\textwidth]{b6.pdf}}
%    % \vspace{-1em}
%    \caption{Example images from the (a) Cifar-10 and (b) Caltech-101 data sets.}
%       \label{obj_rec}
%    \end{figure*}

\subsection{Experimental Settings}

We conduct comparative experiments with both online and offline feature selection methods. The state-of-the-art online feature selection methods include Alpha-investing, OSFS and Grafting. We choose three representative offline feature selection from the filter, embedded and wrapper models, specifically MI (Mutual Information) \cite{hyvarinen2000independence}, LARS (Least Angle Regression) \cite{efron2004least} and GBFS (Gradient Boosted Feature Selection) \cite{xu2014gradient}. The employed evaluation metrics are accuracy and compactness. Compactness is the number of selected features. Accuracy denotes the classification or verification accuracy based on selected feature space. We also report the results based on global feature space as ``Baseline''. According to authors of \cite{wu2013online}, the maximum number of selected features is set to be 50. The parameters in Alpha-investing are set according to \cite{alpha05}. We tune the parameters in Grafting by cross validation. The inter-group selection of our method is implemented by the efficient %efficient
sparse coding method \footnote{http://ai.stanford.edu/$\sim$hllee/softwares/nips06-sparsecoding.htm} with the parameter $\lambda\in [0.1, 0.8]$.

To simulate online group feature selection, we allow the features to flow in by groups. The features in a group are processed individually. For the data sets with natural feature groups, the pre-existing group structure is used. For the data sets with no natural feature groups, we divide the feature space randomly. Specifically, $F = [G_1, \cdots, G_i, \cdots]$ denotes the global feature stream, we split it into several groups randomly. Each feature group $G_i=[f_{(i-1)*k+1}, f_{(i-1)*k+2}, \cdots, f_{i*k}]$ with dimension $k$. In the case that $m$ is less than 100, dimension less than 100,$k$ is set to be half of the global dimension. Otherwise, $k$ is chosen from $\{m/10,m/100,m/200\}$. This experiment can help to test the robustness of OGFS when there is no natural group information. %{\color{blue}As our experiments are implemented in the server, the other tasks on the server has some influence to the time complexity. }

\subsection{Image classification}

We use Cifar-10 \cite{krizhevsky2009learning} and Caltech-101 \cite{fei2007learning} %Cifar-10 \cite{krizhevsky2009learning} and Caltech-101 \cite{fei2007learning} data sets
for image classification. We first introduce %introduced
the data sets in our experiments and then present the experimental results.
%\begin{figure*}
%\centering
% \includegraphics[width=0.9\textwidth]{cifarAcc.pdf}
% \label{cifarAcc}
%  \caption{Recognition accuracy of comparative methods on Cifar-10 data set.}
%\end{figure*}
The Cifar-10 dataset consists of 60,000 images in 10 classes with 6,000 images per class. We randomly select 1,000 images from each class for training and the rest are used for testing.
The Caltech-101 dataset contains 9,144 images from 102 categories (including a background), including animals, vehicles, flowers, etc. There are 31 to 800 images in each category. We take 5, 10, $\cdots$, 30 images per class for training and take 50 images %up to 50 images
per class for testing. In Caltech-101, we extract the SIFT feature of three-layer pyramid. Then, each image is represented by an $\ell_2$ normalized 21 $\times$ 1024-dimensional sparse-coding feature vector. Thus, the feature stream consists of %consists
$F=[G_1, \cdots, G_{21}]^{T}\in \mathbb{R}^{21\times 1024}$, where $G_i\in R^{1024}$ denotes the SIFT descriptor for the whole image if $i=1$, and $G_i\in \mathbb{R}^{1024} (i>1)$ denotes the SIFT descriptor for a local region of the image. As the Cifar-10 dataset contains tiny images with the size of %tiny image with size of
32$\times$32, we extract the the SIFT feature of two-layer pyramid. %represent the image by an $\ell_2$ normalized 5 $\times$1024-dimensional sparse-coding feature vector. 
Then the feature stream consists of %consists
$F=[G_1, \cdots, G_{5}]^{T}\in \mathbb{R}^{5 \times 1024}$. We adopt a %adopts
linear SVM to test the classification performance of the selected feature space. The involved parameter in SVM model is tuned by 5-fold cross-validation. Details %The details
of experimental results are as follows.
\subsubsection{ Cifar-10}

We first explore the individual performance of the two process in OGFS, denoted as OGFS-Intra and OGFS-Inter respectively. Table~\ref{cifarAvg} reports the compactness, accuracy and the time cost for each algorithm on this dataset.

Considering classification accuracy, OGFS-Intra obtains the best overall accuracy with 51.22\% as shown in Table~\ref{cifarAvg}. Grafting performs only after OGFS-Intra with 51.00\%. OGFS-Inter and OGFS reach comparative %comparative
accuracy with 49.54\% and 49.58\%, respectively. Alpha-investing is about 7\% inferior to OGFS-Inter and OGFS, but it still performs better than OSFS. This is possibly because of the constraint on the maximal number of selected features in OSFS. It is demonstrated %demonstrate
that OGFS-Intra can select discriminative features, but leads to redundancy. %but may also include redundant features.
OGFS-Inter can reduce the redundancy. Thus, OGFS achieves %obtain
better accuracy than OGFS-Inter and is a little inferior to OGFS-Intra. The three offline feature selection methods obtain comparative accuracy around 48.00\%. The accuracy of Baseline is the best with 54.40\%. We can observe that the accuracy gap between our method and Baseline is the least.

In terms of compactness, %OSFS achieves the best compactness but with the sacrifice of classification accuracy. Alpha-investing is only inferior to OSFS with much fewer features compared with others.
as shown in Table~\ref{cifarAvg}, OSFS selects only 50 features. OGFS-Intra selects the largest number of features (5,111), is similar with Grafting (4,945). This is because OGFS-Inter uses a sparse model which leads to a relatively small size of feature space. GBFS obtains the least number of features among offline feature selection methods with 1,694, but our OGFS is comparative with 1,990 features. To guarantee the classification performance of MI, MI selects the same size of features as LARS (2,723).  %The previously selected features by Alpha-investing dominate the following selection by updating the threshold $\alpha$.

In terms of time complexity, OGFS-Intra obtains the highest efficiency with only %with only
3.53 seconds, while others require %needs
hundreds or thousands of seconds. This is because %It is because
OGFS-Intra is linear with the number of features as we discuss %discusses
in Section~\ref{ogfs33}. The inter-group selection needs less than 150 seconds, which is much %much
faster than Alpha-investing, Grafting and OSFS. This is because the time cost of OSFS is exponential with the number of desired features. In order to simulate the online situation, all the online feature selection methods tend to spend more time in feature transformation. The time complexity of the filter method MI is 8.62 seconds, much faster than other offline methods, LARS (121.18) and GBFS (281.17). However, our OGFS-Intra is even more efficient with only 3.53 seconds. This is the benefits of our criterion defined in intra-group selection.

\begin{figure}
\centering
   \subfloat[]{ %0.42
        \includegraphics[width=0.43\textwidth]{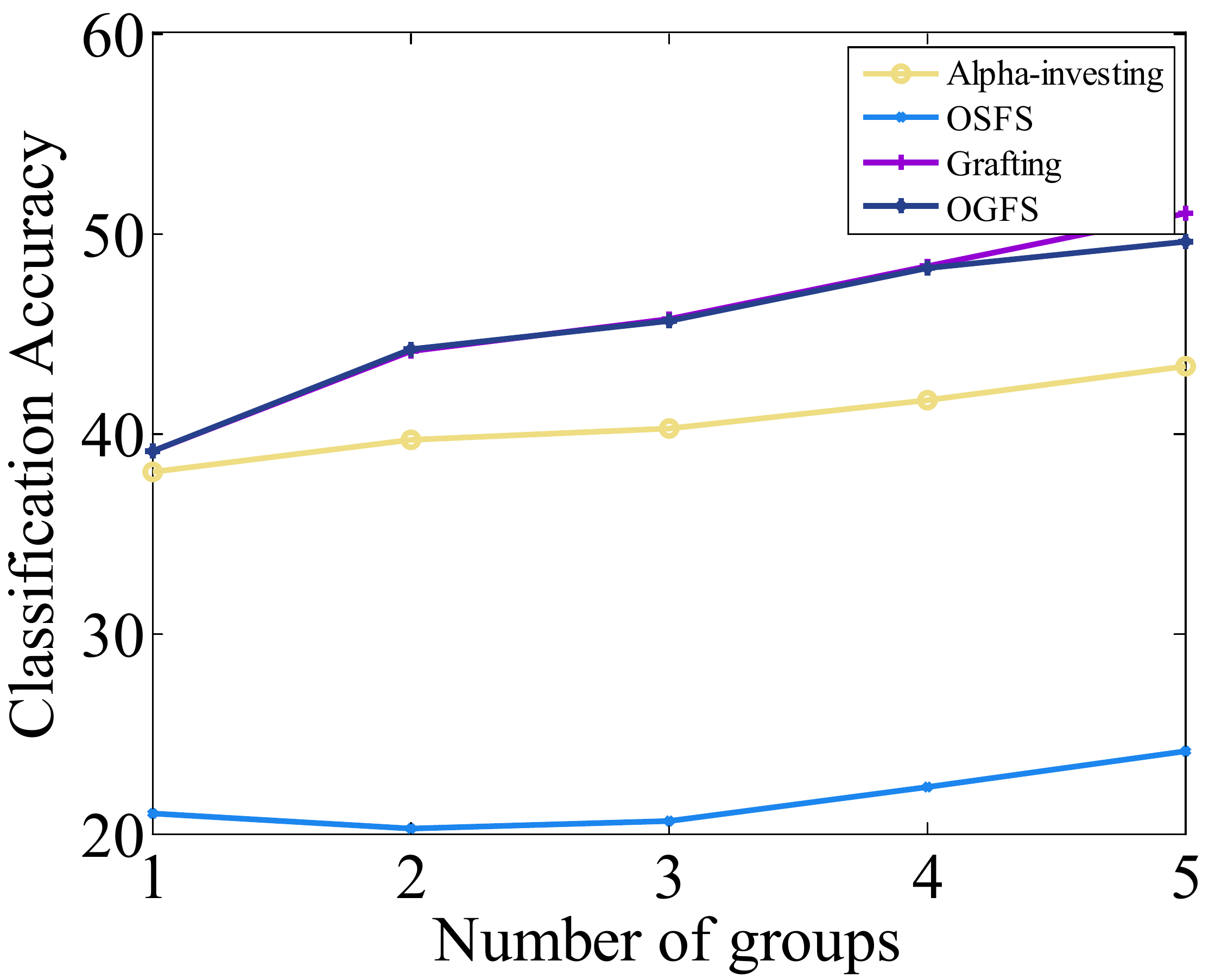}}
%       ~~~~~~  \hfill
     \subfloat[]{ %0.45
        \includegraphics[width=0.46\textwidth]{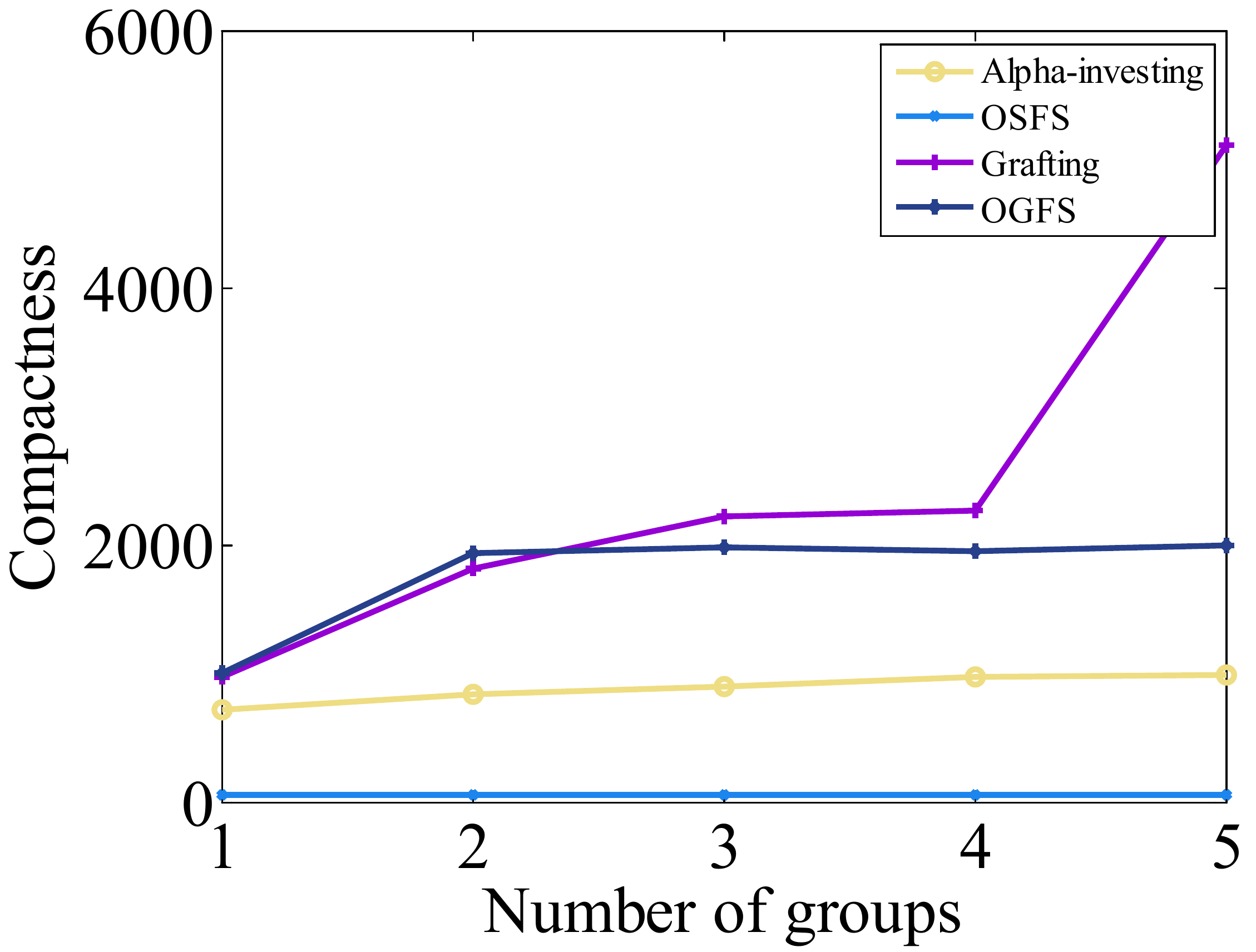}}

    \caption{The performance of online feature selection vs. feature groups on the Cifar-10 dataset.}
	\label{cifar_acc_dim}
	\vspace{-6.5em}
\end{figure}

Since we studied the online feature stream with groups, we examine the performance of online feature selection methods in response to increasing groups in Figure~\ref{cifar_acc_dim}. Generally, with the arrival of more groups, the compactness increases and the classification accuracy improves. But the improvement is not obvious for Alpha-investing and OSFS. Grafting and our method obtains the best accuracy. But our method obtains a better compactness. Actually, when the number of groups increases to 2, the compactness of our method remains stable. This is because the complementary effects of the two stages of OGFS. The OGFS-Intra selects the most discriminative features, and OGFS-Inter achieves the optimal subset.

To sum up, benefit from group information, OGFS favors a good trade-off between the accuracy and compactness. The time complexity show that the combination of the two stages (OGFS-Intra and OGFS-Inter) is reasonable and applicable for real-world applications. Thus, in the following experiments, we only compare our OGFS algorithm with other comparative algorithms. %our algorithm OGFS with other comparative algorithms.

\begin{table}[!t] %(table*)
\centering
\caption{Image classification results on the Cifar-10 dataset.}
\label{cifarAvg}
\centering
\begin{tabular}{|l| r| l| r| }
\hline
%\multirow{2}*{Training images} & \multicolumn{2}{c|}{Alpha-investing} & \multicolumn{2}{c|}{Fast-OSFS} & \multicolumn{2}{c|}{Grafting }& \multicolumn{2}{c|}{OGFS}\\
Method &$\#$dim. &accu. &time(s)\\ \hline
Alpha-investing	&$\textbf{979}$      &43.31 	&3228.82  \\ \hline
OSFS	&50	&24.07	&45625.17 	 \\ \hline
Grafting	&4945	&51.00 &4562.88\\ \hline
OGFS-Intra	&5111	&51.22 &$\textbf{3.53}$ \\ \hline
OGFS-Inter	&1991 &49.54 &142.98  \\ \hline
OGFS	&1990	&49.58 &142.09 \\ \hline
MI &2723 & 49.43 &8.62\\ \hline
LARS &2723 &47.24 &121.18\\ \hline
GBFS &1694 &48.54 &281.17\\ \hline
Baseline &5120 &\textbf{54.40} &-\\ \hline
\end{tabular}
\end{table}

\begin{figure*}
\centering
% 0.32\includegraphics[width=1.0\textwidth]{caltech_group_acc.pdf}
 \subfloat[5 Training.]{
        \includegraphics[width=0.3\textwidth]{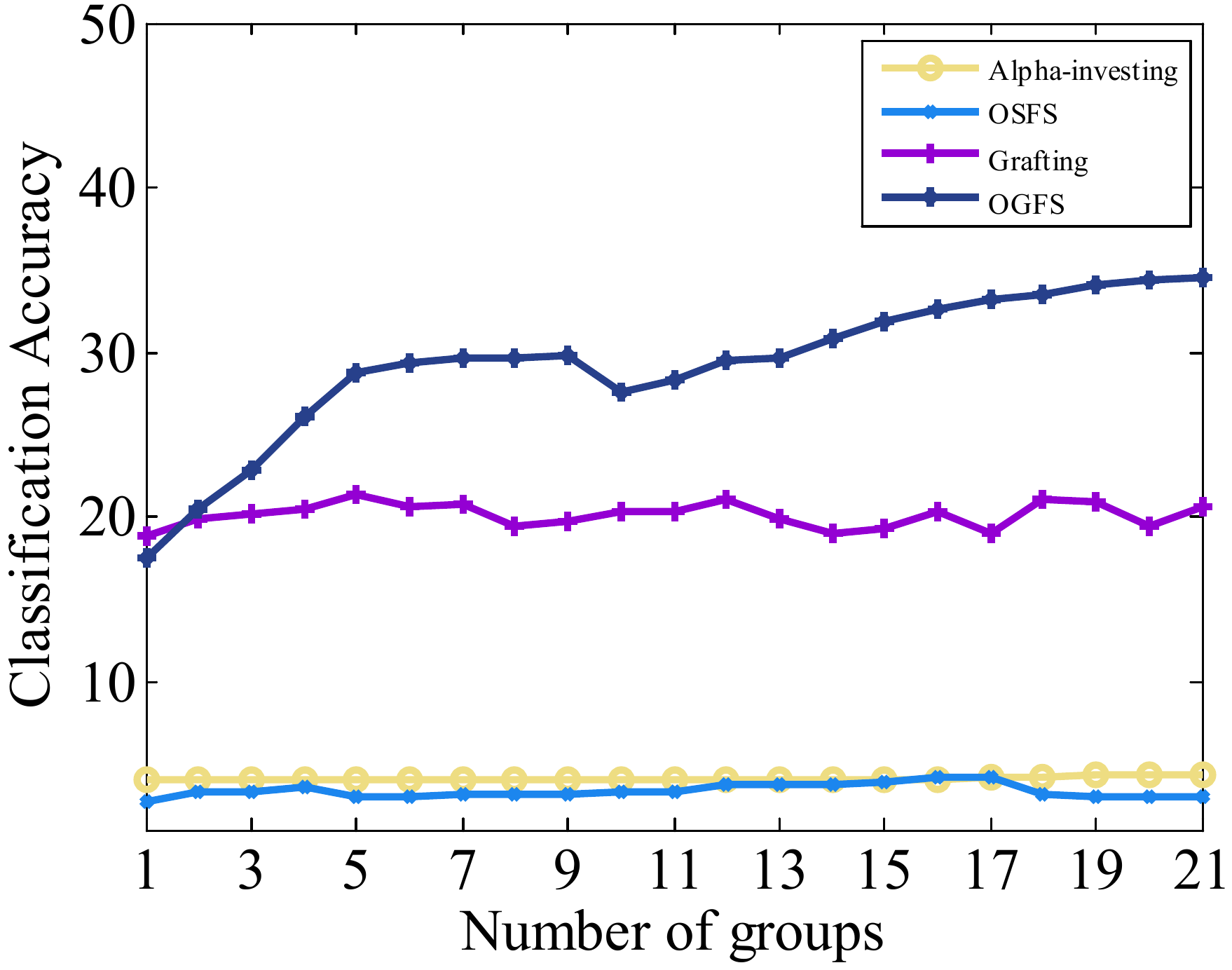}}~~
%         \hfill
 \subfloat[10 Training.]{
     \includegraphics[width=0.3\textwidth]{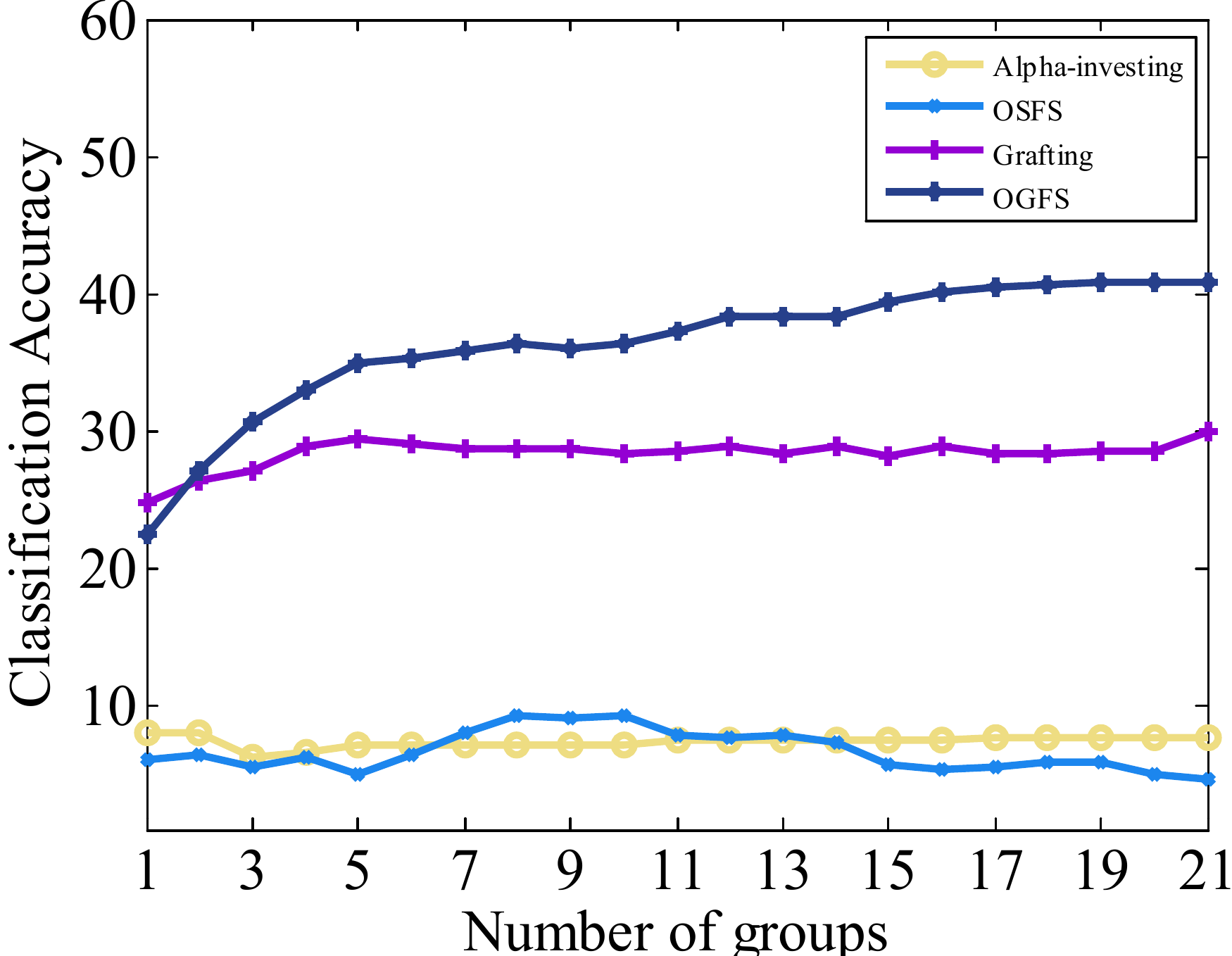}}~~
 \subfloat[15 Training.]{
     \includegraphics[width=0.3\textwidth]{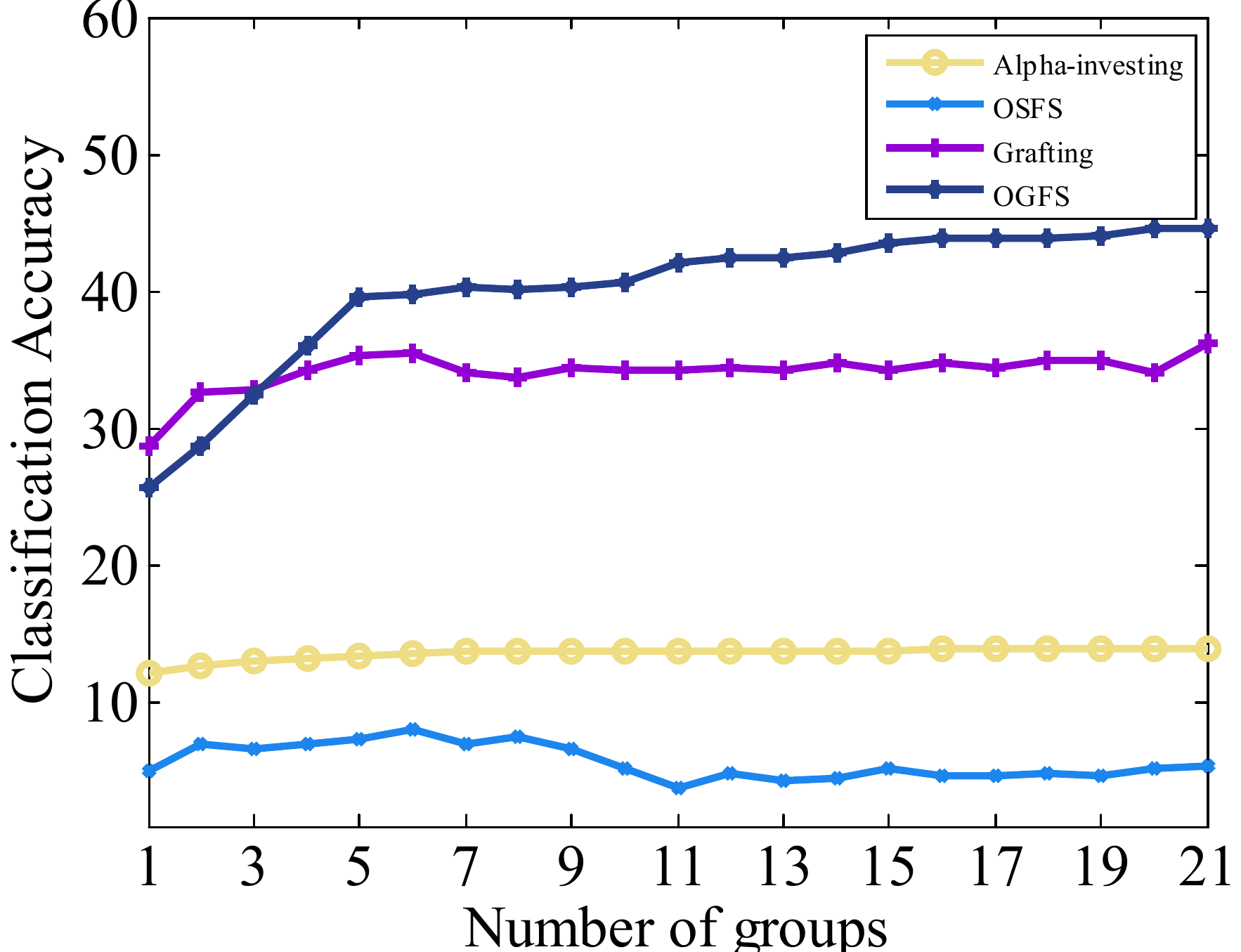}}\\
 \subfloat[5 Training.]{
     \includegraphics[width=0.3\textwidth]{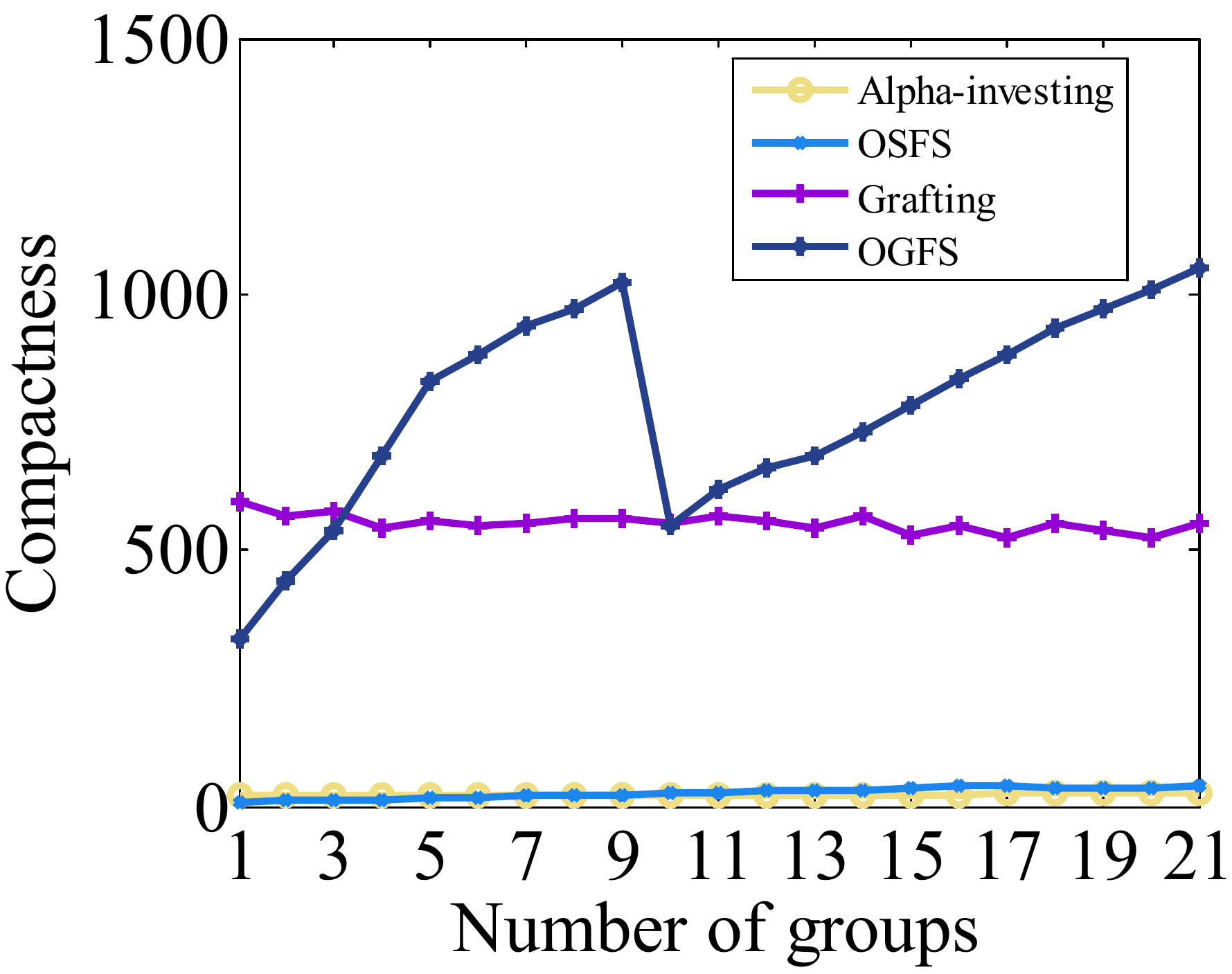}}~~
 \subfloat[10 Training.]{
    \includegraphics[width=0.3\textwidth]{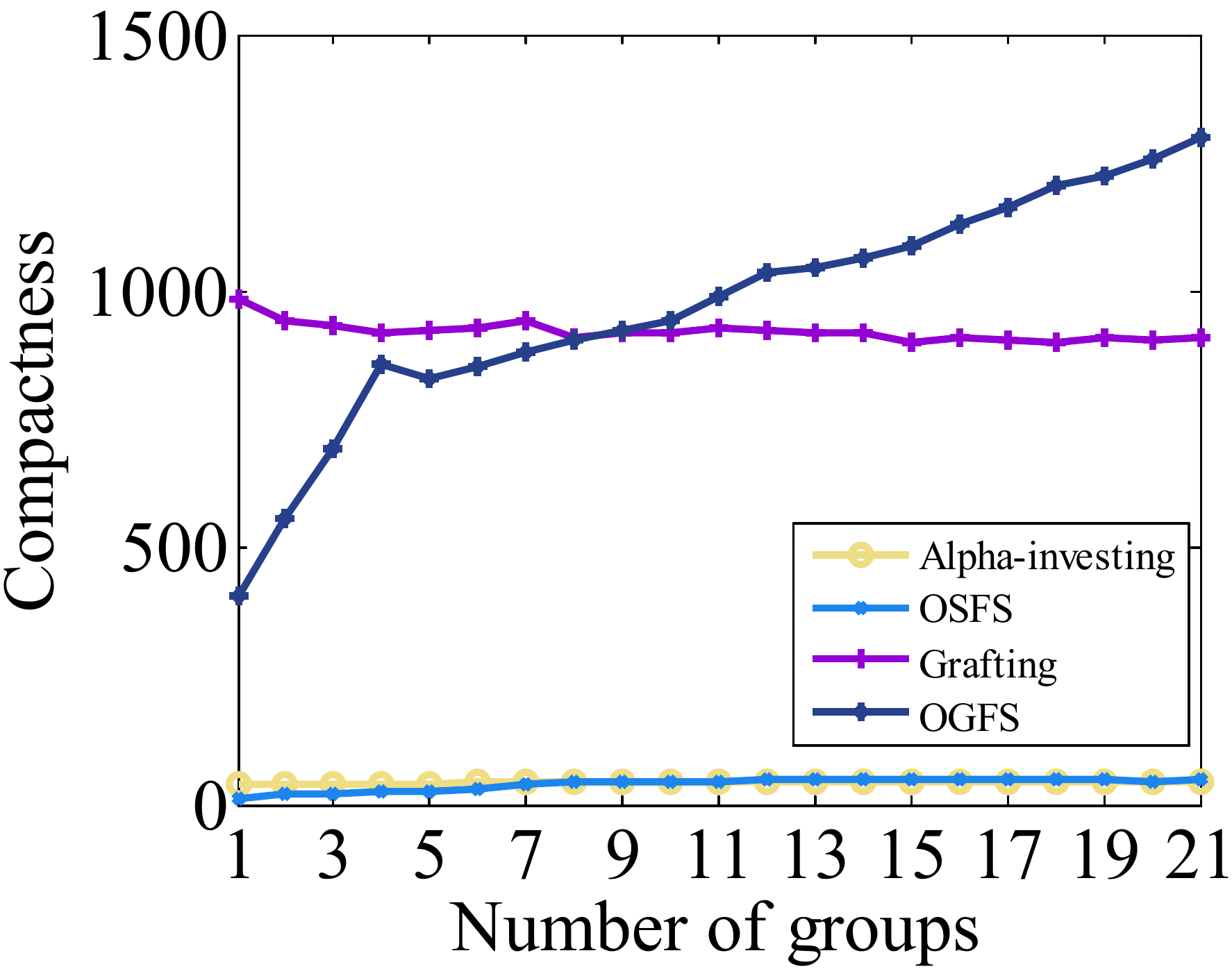}}~~
 \subfloat[15 Training.]{
    \includegraphics[width=0.3\textwidth]{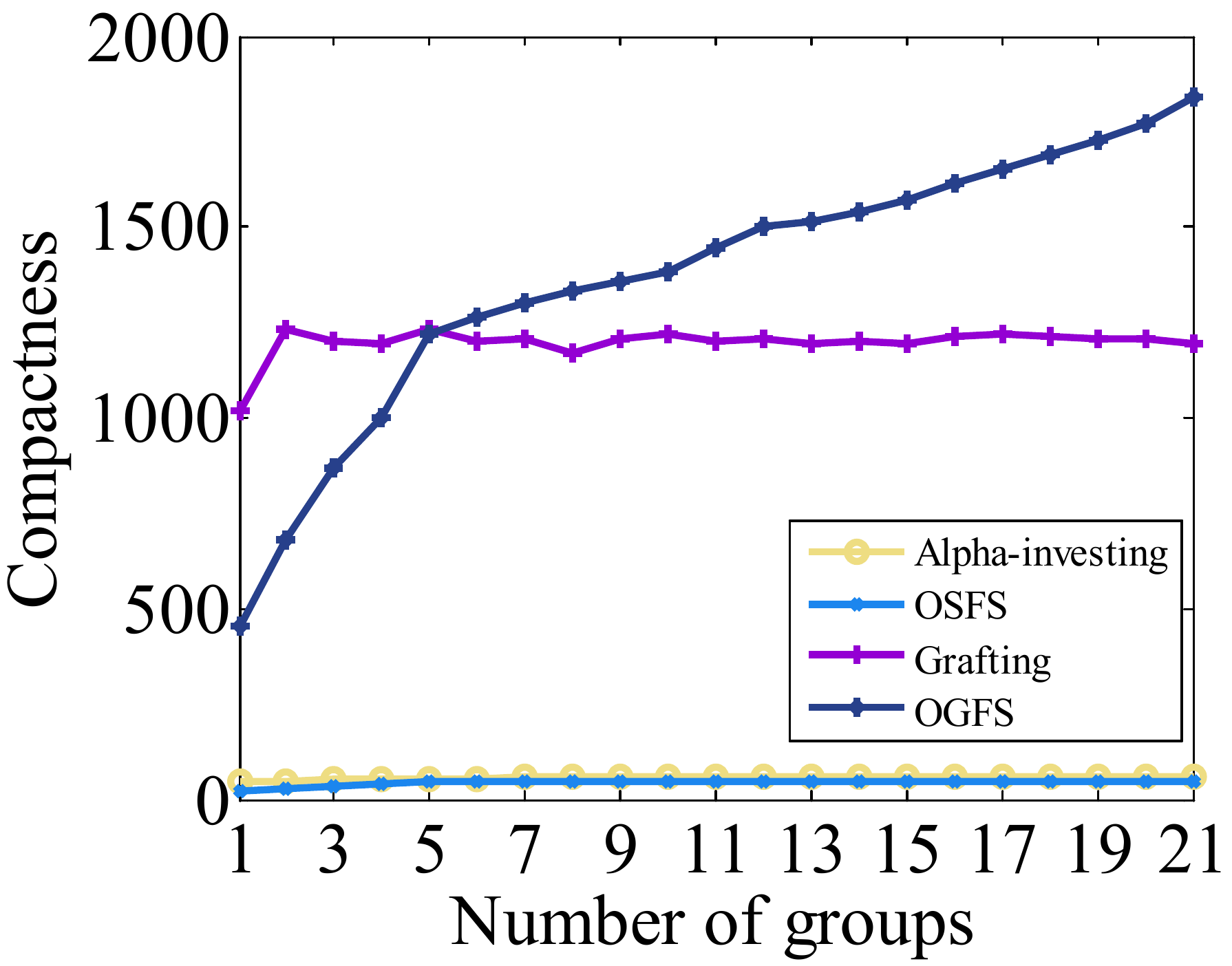}}
   \caption{\small{The performance of online feature
   selection vs. the number of feature groups on the Caltech-101 dataset.}}
   \label{cal_group_acc}
\end{figure*}

%\begin{figure*}
% \includegraphics[width=1.0\textwidth]{caltech_group_dim.pdf}
%   \caption{Compactness evaluation of online feature selection in the entire online learning process on the Caltech-101 data sets with various training images per class.}
%   \label{cal_group_dim}
%\end{figure*}

\begin{table*}[!t] %(table*)
\centering
\caption{Image classification results on the Caltech-101 dataset with online feature selection methods.}
\label{caltechRes}
\centering
 \resizebox{\textwidth}{!}{  
\begin{tabular}{|c|r | r|r|r|r| r| r| r |r| r |r|r| } %{|c| c| c| c| c| c |c| c |c|}
\hline
\multirow{2}*{Train} & \multicolumn{3}{c|}{Alpha-investing} & \multicolumn{3}{c|}{OSFS} & \multicolumn{3}{c|}{Grafting }& \multicolumn{3}{c|}{OGFS} \\
\cline{2-13} &$\#$dim.	&accu.&time(s)	&$\#$dim.	&accu.&time(s)	&$\#$dim.	&accu.&time(s) &$\#$dim.	&accu.&time(s)\\ \hline
5	&$\textbf{25}$      &4.24 &$\textbf{12.19}$	&38	&3.02 &201.80 &553	&20.61 &569.04	&1,051	&$\textbf{34.54}$ &140.48   \\ \hline
10	&$\textbf{46}$	&7.04 &$\textbf{30.56}$	&50	&4.62 &2312.6 &1,258 &29.98 &1976.82	&1,302 &$\textbf{40.92}$	&192.06 \\ \hline
15	&$\textbf{60}$	&12.23 &$\textbf{55.64}$ &50	&4.89 &5971.8   &1,196	&36.23 &754.92&	1,842&$\textbf{44.55}$ &173.70  \\ \hline
20	&$\textbf{79}$	&15.20 &$\textbf{113.33}$ &50	&5.76 &1203.3 &1,390	&38.38&1008.79&1,495	&$\textbf{48.98}$ &237.12  \\ \hline
25	&$\textbf{118}$ &20.14 &250.81 &50	&6.39	&1405.8 &1,528 &41.44&2024.70&	1,856	&$\textbf{52.67}$ &$\textbf{220.39}$    \\ \hline
30	&$\textbf{109}$	&20.58 &$\textbf{266.49}$ &50	&5.48 &2137.6 &1,641 &45.21 &2470.00&1,782	&$\textbf{52.05}$  &327.54  \\ \hline
\end{tabular}
}
\end{table*}

\begin{table*}[!t] %(table*)
\centering
\caption{Image classification results on the Caltech-101 dataset with offline feature selection methods.}
\label{calFullRes}
\centering
 \resizebox{\textwidth}{!}{  
\begin{tabular}{|c|r | r|r|r|r| r| r| r |r| r |r|r|  } %{|c| c| c| c| c| c |c| c |c|}
\hline
\multirow{2}*{Train} & \multicolumn{3}{c|}{MI} & \multicolumn{3}{c|}{LARS} & \multicolumn{3}{c|}{GBFS}& \multicolumn{2}{c|}{Baseline} \\
\cline{2-12} &$\#$dim.	&accu.&time(s)	&$\#$dim.	&accu.&time(s)	&$\#$dim.	&accu.&time(s) &$\#$dim.	&accu.\\ \hline
5	&500	&18.36 &12.89 &502	&16.79 &10.42	&\textbf{318}	&16.00 &641.21 &21,504 &\textbf{39.74}  \\ \hline
10	&1,001	&29.08 &16.14 &1,001 &30.58 &41.25	&\textbf{734} &27.49	&364.45 &21,504 &\textbf{49.02} \\ \hline
15	&1,511	&35.47 &20.29   &1,511	&36.35 &90.42 &\textbf{1,047}&34.23 &526.60 &21,504 &\textbf{54.95} \\ \hline
20  &2,014	&42.09 &24.74 &2,014	&41.88 &160.32 &\textbf{1,372}	&39.43 &699.48 &21,504 &\textbf{57.93} \\ \hline
25  &2,509	&48.22	&28.85 &2,509 &47.12 &257.54 &\textbf{1,674}	&41.44 &876.44   &21,504 &\textbf{62.24} \\ \hline
30  &3,000	&52.15 &32.92 &3,000 &51.35 &381.63 &\textbf{1,966}	&45.68  &1065.29 &21,504 &\textbf{64.51} \\ \hline
\end{tabular}}
\end{table*}

\subsubsection{ Caltech-101}
We report the average accuracy over 101 classes. Detailed results are shown in Table~\ref{caltechRes}. %£¬
It can be seen that OGFS gives the leading classification accuracy in all the cases. Specifically, OGFS gains 30\% over Alpha-investing. The performance of Grafting improves by the increase of training samples, but is still inferior to our method. The accuracy of OGFS is about 6$\sim$13\% higher than Grafting. For example, in the case of 30 training images, Grafting reaches the accuracy of 45.21\% while our method is 52.05\%. In the case of 25 training images, the accuracy of other methods are all below 45\% while OGFS reaches 52.67\%. 

In terms of compactness, Alpha-investing achieves the best performance. In the case of 20 training images per class, Alpha-investing obtains the compactness with 79 features, much better than the comparative methods, such as Grafting (1,390) and OGFS (1,495). However, Alpha-investing only achieves the accuracy of 15.20\%, much lower than Grafting (38.38\%) and OGFS (48.98\%). It implies that the reevaluation of the features is necessary. This also confirms that the correlation among the features is important.

In terms of time complexity, the time complexity of all methods increases by the increase of training samples. %Except in several cases, such as the training images increases from 10 to 15, the time cost of Grafting decreases from 1976.82 to 754.92 seconds. This is because though the processing time of each feature increases, the global time complexity also depends on the number of previous selected features. In summary, 
Alpha-investing is the most efficient in most cases. However, in the case of 25 training images, OGFS is 30 seconds faster. OSFS and Grafting achieve similar computational efficiency which varies from 190.0 to 2500.0 seconds. Thus, in summary, OGFS could obtain the most discriminative feature space  within acceptable time cost.

Table~\ref{calFullRes} reports the offline feature selection methods and Baseline. Baseline obtains the best accuracy but with huge feature space. In the case with less than 20 training samples, the offline feature selection methods obtain comparative accuracy with less than 3.00\% variation. With the increase of training samples, MI enjoys a great improvement in accuracy. In the case of 30 training samples, MI obtains the best accuracy with 52.15\%, better than LARS (51.35\%) and GBFS (45.68\%). OGFS is comparative with 52.05\%. The results demonstrate that OGFS is superior than offline feature selection methods in the real-world image classification task.

We investigate the influence of increasing feature groups. The classification results based on each group is plot in Figure~\ref{cal_group_acc}. We can also observe that with the increase of feature groups, OGFS enjoys an improvement in accuracy. For instance, as shown in Figure~\ref{cal_group_acc}(b), OGFS obtains much better accuracy than Grafting when there are 3 groups. But when the feature group reaches to 5, the performance of most methods keep steady. The compactness of our method changes with the increase of groups while others remain stable. It demonstrates the efficacy of online group feature selection. The features extraction is expensive and time consuming. If the model based on existing feature space reaches predefined performance, the further feature extraction is not necessary.
\begin{figure}
\centering
 \includegraphics[width=0.3\textwidth]{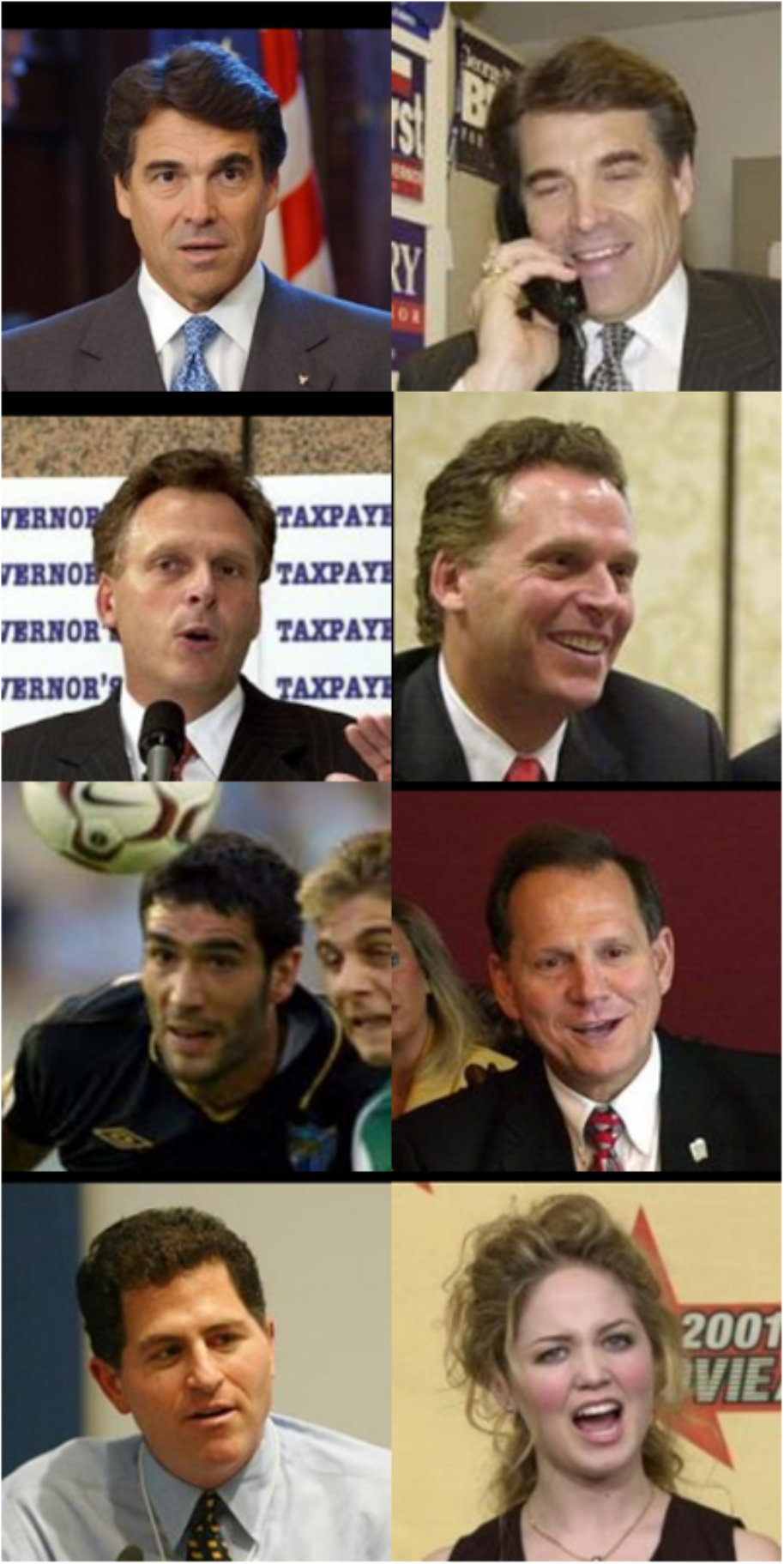}
   \caption{Pair of samples from the LFW dataset. The first two rows are correctly matched pairs and the last two rows are mismatched pairs.}
   \label{lfw_Sample}
\end{figure}

\begin{table*}[!t] %(table*)
\centering
\caption{Face verification results on the LFW dataset}
\centering
\label{lfwRes}
\begin{tabular} {|c|r | r| r| r| r |r| r r|} %{|c| c| c| c|c| c |c|c |c|}
\hline
\multirow{2}*{Fold} & \multicolumn{2}{c|}{Alpha-investing} &  \multicolumn{2}{c|}{OSFS}& \multicolumn{2}{c|}{Grafting}& \multicolumn{2}{c|}{OGFS}\\
\cline{2-9} &$\#$dim.	&accu.		&$\#$dim.	&accu. &$\#$dim.	&accu.&$\#$dim.	&accu.\\ \hline
1	&$\textbf{1}$      &52.67 	 &50	&66.17 	&3,963	&77.00 &1,132 &$\textbf{79.50}$\\ \hline
2 &$\textbf{1}$      &52.67	&50	&67.50 &3,965	&77.50 &1,619 &$\textbf{82.33}$ \\ \hline
3 &$\textbf{2}$      &54.67&50	&66.83   &3,867	&77.33 &1,915 &$\textbf{81.17}$   \\ \hline
4 &$\textbf{1}$      &52.67 &50	&62.67 &3,961 &77.17   &1,602 &$\textbf{81.17}$   \\ \hline
5 &$\textbf{1}$      &52.67  &50	&65.50 &4,004 &76.50 &1,590 &$\textbf{81.00}$\\ \hline
6 &$\textbf{1}$      &52.67  &50	&64.50  &3,825 &77.33&1,695 &$\textbf{81.33}$\\ \hline
7  &$\textbf{1}$      &52.67 &50	&66.17   &3,674 &77.33 &1,536 &$\textbf{80.83}$ \\ \hline
8 &$\textbf{1}$      &52.67 &50	&69.50   &3,825 &77.50  &1,411 &$\textbf{80.67}$\\ \hline
9 &$\textbf{1}$      &52.67 &50	&65.50  &3,831 &77.17&1,716 &$\textbf{80.00}$ \\ \hline
10 &$\textbf{2}$      &54.67 &50	&66.67  &3,844 &76.83 &1,338 &$\textbf{81.17}$ \\ \hline
average &$\textbf{1}$ &53.07$\pm$0.84 &50 &65.70$\pm$1.36 &3,876 &77.17$\pm$0.31 &1,555	&$\textbf{80.92}\pm$0.77\\ \hline

\end{tabular}
\end{table*}%(table*)??

\begin{table}[!t] %(table*)
\centering
\caption{Face verification results with feature selection on the LFW dataset}
\label{lfwAvg}
\centering
\begin{tabular}{|l| r|r|  }
\hline
%\multirow{2}*{Training images} & \multicolumn{2}{c|}{Alpha-investing} & \multicolumn{2}{c|}{Fast-OSFS} & \multicolumn{2}{c|}{Grafting }& \multicolumn{2}{c|}{OGFS}\\
Method &$\#$dim. &accu. \\ \hline
Alpha-investing & \textbf{1} & 52.67\\ \hline
OSFS &50 & 62.67 \\ \hline
Grafting & 3,961 &77.17 \\ \hline
OGFS & 1,602 &\textbf{81.17} \\ \hline
MI &5,000 & 76.50\\ \hline
LARS &4,073 &77.17\\ \hline
GBFS &68 &67.50 \\ \hline
Baseline &127,440 &76.83 \\ \hline
\end{tabular}
 \vspace{-0.1in}
\end{table}

\begin{figure}
\centering
   \subfloat[]{
        \includegraphics[width=0.45\textwidth]{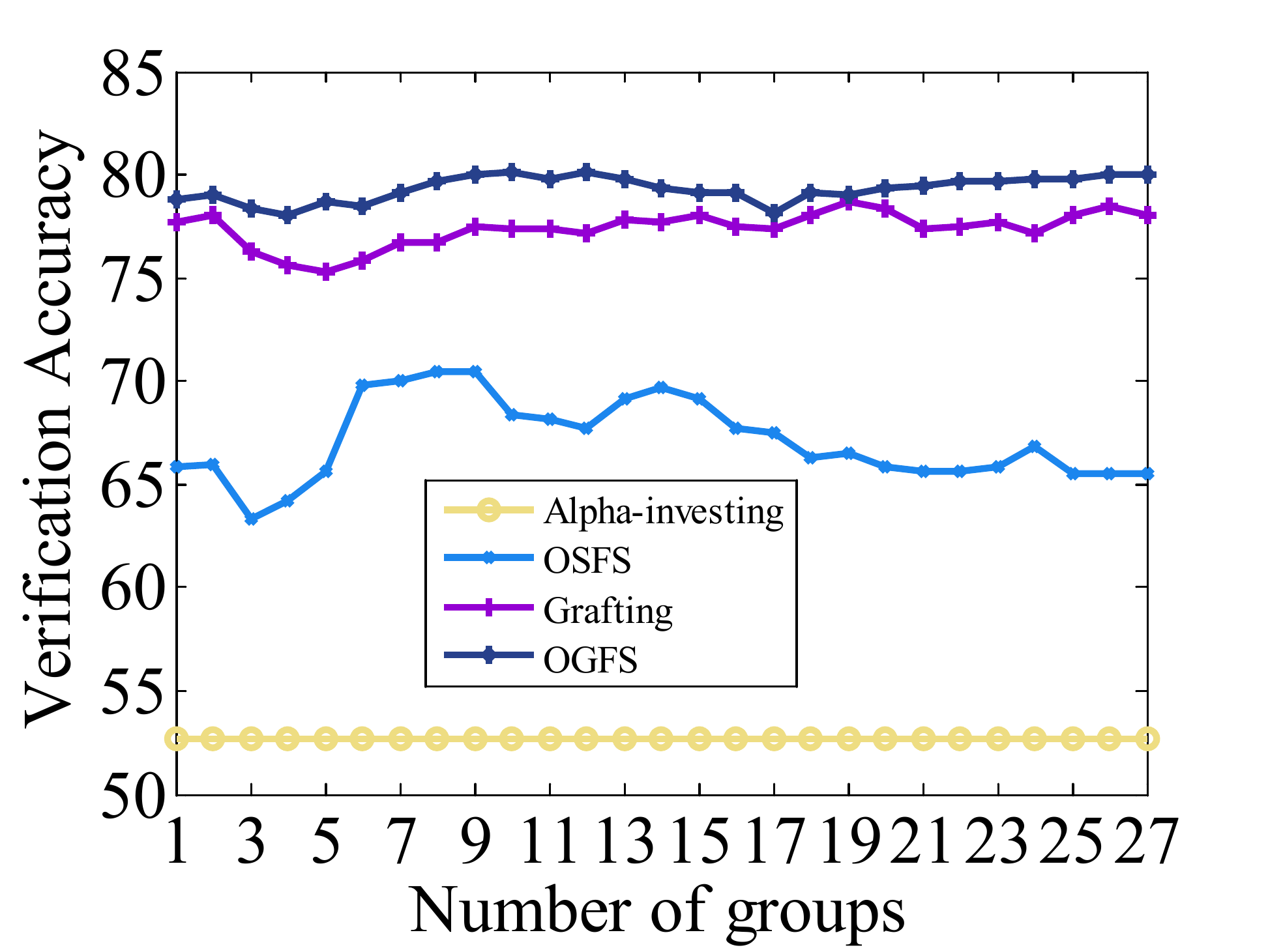}}
%  0.42    ~~~~~~   \hfill
     \subfloat[]{
        \includegraphics[width=0.45\textwidth]{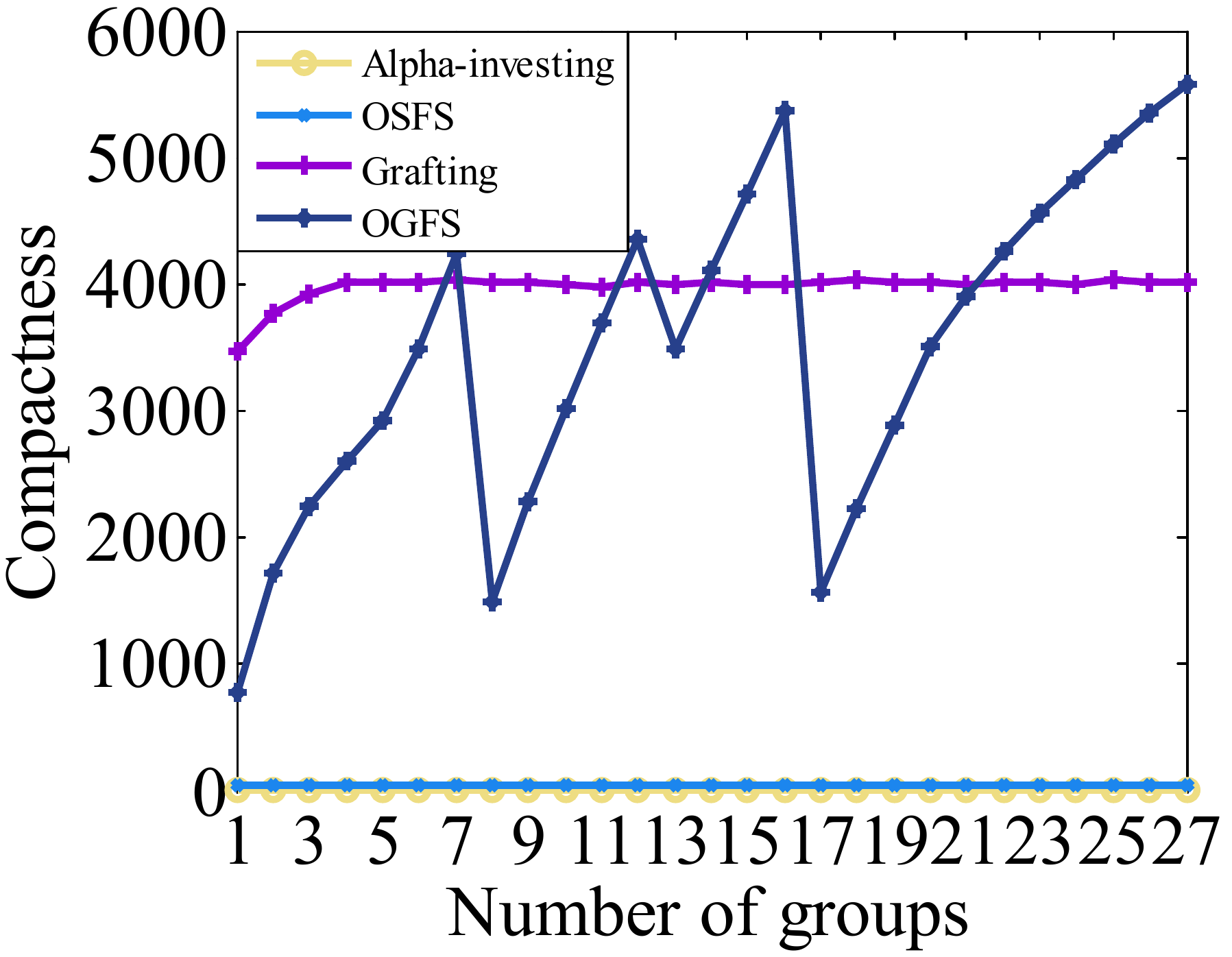}}

    \caption{The performance of online
    feature selection vs. feature groups on the LFW dataset.}
	\label{lfw_acc_dim}
	\vspace{-1em}
\end{figure}

\subsection{Face Verification}
The LFW dataset is collected for unconstrained face recognition~\cite{huang2007labeled}. It contains 13,233 images from 5,749 identities. In this dataset, there are over 1,680 identities that have two or more images, and 4,069 identities that have just one single image. The images are captured under daily conditions with variations in pose, expression, age, lighting and so on. Figure~\ref{lfw_Sample} lists some samples in the dataset.

We extract image patches at 27 landmarks in 5 scales. The patch size is fixed to 40$\times$ 40 in all scales. Each patch is divided into 4$\times$4 non-overlapping cells. For each cell, we extract the 58-dimensional LBP descriptor. Then each image is represented by a feature vector with dimension 27$\times$5$\times$4$\times$4$\times$58. We set the feature space of each landmark as a group. Then, the feature stream consists of $F=[G_1, \cdots, G_{27}]^{T}\in \mathbb{R}^{27\times4640}$, where $G_i\in R^{1024}(i\in[1,27])$ denotes the LBP descriptor for a landmark. The dataset is divided into ten folds. We test the performance on each selected feature space in a leave-one-out cross validation scheme. In each experiment, nine folds are combined to form a training set, with the tenth subset used for testing. We verify whether each pair belongs to the same subject by Euclidean distance.
Table~\ref{lfwRes} lists the details of the compactness and the verification accuracy on selected feature spaces of each fold. %Details of the compactness and the recognition accuracy on selected feature spaces of each fold are seen in Table~\ref{lfwRes}.
%The bottom line is the average accuracy with standard variance.

As shown in Table~\ref{lfwRes}, OGFS is over 20\% higher than Alpha-investing in all cases. Alpha-investing selects only 2 or 4 features. The indices of selected features are among \{1, 2, 3, 4, 5, 396\}. This is because the previously selected features are never reevaluated. It confirms the importance of reevaluating collected features. In general, OSFS achieves the accuracy about 66.00\% with only 50 features, much higher than Alpha-investing (about 53.00\%), but still inferior than OGFS (80\%). OGFS also outperforms Grafting in both accuracy and compactness. For instance, on the 3rd fold, Grafting achieves 77.33\% accuracy with 3,857 features, while OGFS achieves 82.33\% with 1,619 features.

In terms of time complexity, Alpha-investing still obtains the highest efficiency with 137.23 seconds in average. This is because the time complexity of Alpha-investing is linear. OSFS is only inferior to Alpha-investing with 2470.57 seconds. Grafting is the slowest with over 76,000 seconds, much slower than our method 4752.93 seconds. This is because the time complexity of OSFS, Grafting and OGFS are all related to the selected number of features, while Alpha-investing is only correlated to the procession of each dimension of feature. The time complexity of our method is acceptable.

From Table~\ref{lfwRes}, the variance of the 10 splits of data is small. Therefore, we use the 5th fold of data to test the offline feature selection methods. Complementary results are shown in Table~\ref{lfwAvg}. From Table~\ref{lfwAvg}, OGFS obtains the best accuracy with 81.17\%, even better than Baseline with 76.83\%. It demonstrates the necessary of feature selection in face verification. MI and LARS reach similar accuracy with 76.50\% and 77.17\%. Grafting also obtains better accuracy than offline methods. The encouraging results show the superiority performance of online feature selection methods.

\begin{figure}
\centering
 \includegraphics[width=0.4\textwidth]{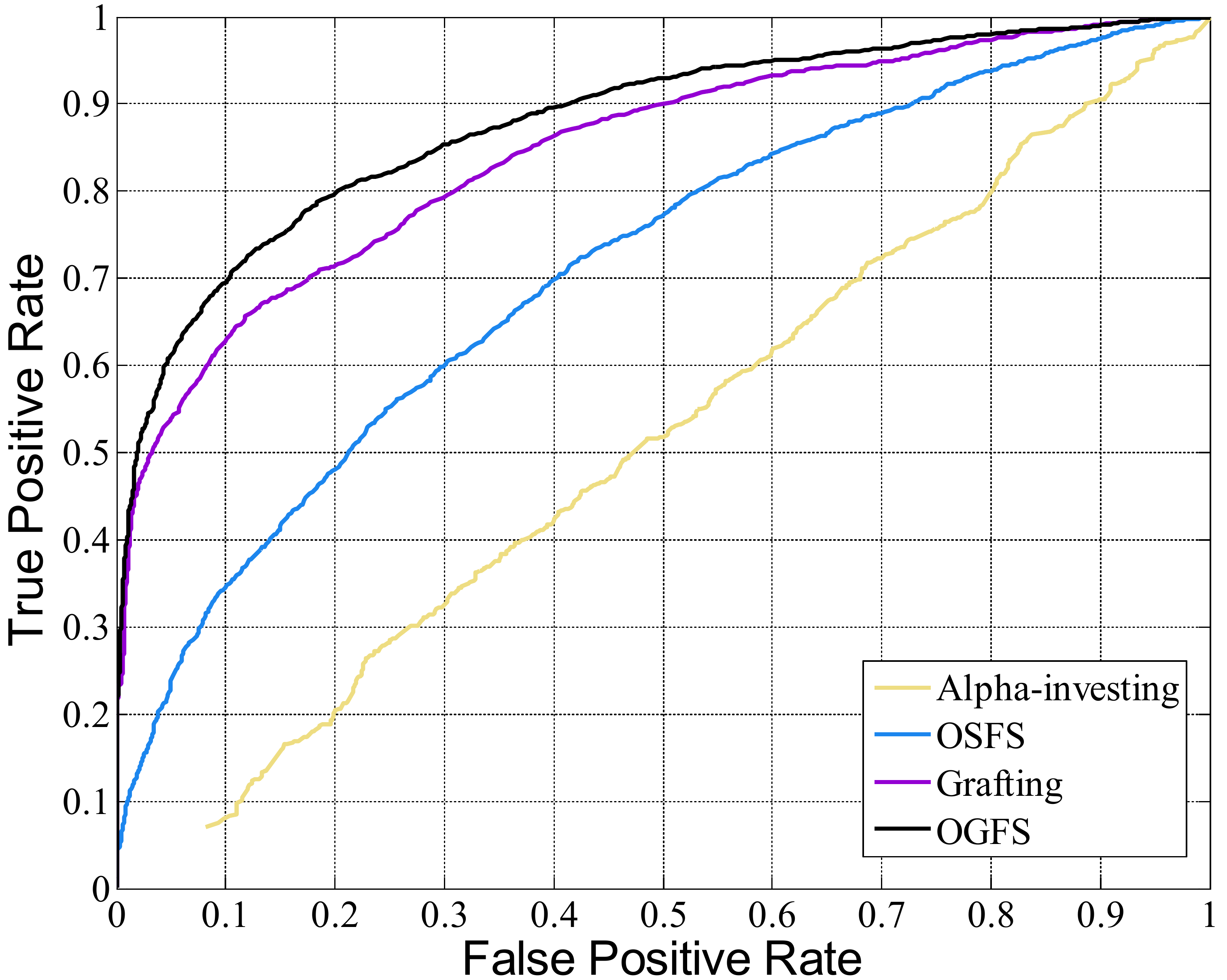}
   \caption{The face verification results on LFW dataset. }
   \label{lfwAcc}
\end{figure}

Figure~\ref{lfwAcc} represents the Receiver Operating Characteristic (ROC) curves of the four methods, from which we can also clearly see the superiority of the proposed OGFS method.

Figure~\ref{lfw_acc_dim} illustrates the performance of online feature selection methods in response to increasing groups. Alpha-investing remains stable in terms of both accuracy and compactness. As the number of groups increases, OSFS and Grafting gain stable compactness. But sometimes the accuracy also decreases. This implies that more features may include redundant or irrelevant information. The results demonstrate that the framework of online feature selection is suitable for the large-scale real-world application.

\begin{table}[!t]
\centering
\caption{Description of the UCI datasets}
\label{bench}
\centering
\begin{tabular}{|l| c| r|r| }
\hline
%Data Set	&$\#$classes	&\multicolumn{2}{c|}{$\#$instances} & $\#$dim.	& $\#$groups \\ \hline& $\#$groups
Data Set	&$\#$classes	&$\#$instances & $\#$dim.	 \\ \hline
Wdbc	&2	&569 & 31	 \\ \hline
Ionosphere	&2	&351	&34	\\ \hline
Spectf	&2	&267 &	44\\ \hline
Spambase	&2&4,601		&57\\ \hline
Colon	&2	& 62	&2,000\\ \hline
Prostate	&2	& 102	&6,033	\\ \hline
Leukemia	&2&	72	&7,129	\\ \hline
Lungcancer	&2	&181	&12,533	\\ \hline
%\multirow{2}*{Soccer}&	\multirow{2}*{7} &	$\#$train	& $\#$test	&\multirow{2}*{182 }  &\multirow{2}*{3}	\\ \cline{3-4}
%& & 196 & 84 & & \\ \hline
%Flower-17&	17&	680	&340	&182	&3 \\ \hline
%15 Scenes&	15&	1500	&2,985	&21,504	&21 \\ \hline
\end{tabular}
\end{table}

\begin{table*}[!t] %(table*)
\centering
\caption{Experimental results of online feature selection methods on benchmark data sets.}% by (a) Alpha-investing, (b) Fast-OSFS, (c) Baseline, and (d) OGFS.}
\label{resultBench}
\centering
 \resizebox{\textwidth}{!}{  
\begin{tabular}{|l|r | r| r|r| r |r| r| r | r | r | r | r|} %{|c| c| c| c| c| c |c| c |c|}
\hline
\multirow{3}*{Data Set} & \multicolumn{3}{c|}{Alpha-investing} & \multicolumn{3}{c|}{OSFS} & \multicolumn{3}{c|}{Grafting } & \multicolumn{3}{c|}{OGFS}\\
\cline{2-13} &$\#$dim.	&accu.	&time(s)            &$\#$dim.	&accu. &time(s) &$\#$dim.	&accu. &time(s) &$\#$dim.	&accu. &tim(s)\\ \hline
Wdbc    	&19    	    &$\textbf{96.84}$	&$\textbf{0.010}$         &$\textbf{11}$	&94.39  &0.182      &19	&95.79   &7.305	 &19  	    &95.26 &0.461 \\ \hline
Ionosphere	&$\textbf{2}$	&91.76  &$\textbf{0.004}$         &9	        &$\textbf{92.60}$   &0.029     &32	&91.76 &0.300    &23  	    &91.47 &0.018\\ \hline
Spectf   	&$\textbf{2}$      	&79.50    &$\textbf{0.002}$        &4	&79.06    &0.034   &44   &80.56  &0.510      &33        &$\textbf{81.27}$ &0.019\\ \hline
Spambase	&$\textbf{42}$        &91.02     &0.200       &84     	&$\textbf{94.07}$ &0.551 &55	&92.28 &0.761 &46	&93.09 &$\textbf{0.047}$\\ \hline
Colon     	&$\textbf{4}$	&79.76   &$\textbf{0.127}$     &$\textbf{4}$	&85.95	  &33.855   &26   &84.26  &3.901     &74	    &$\textbf{90.47}$  &2.033  \\ \hline
Prostate	&8	&97.09    &$\textbf{0.633}$      &$\textbf{5}$	        &91.09   &2.903      &17	&93.53  &9.330     &102	    &$\textbf{98.00}$  &13.724\\ \hline
Leukemia	&6	&98.75  &$\textbf{0.731}$      &$\textbf{5}$	        &94.46    &3.913    &13	&94.53  &5.895     &91	    &$\textbf{100.0}$ &9.132 \\ \hline
Lungcancer	&$\textbf{4}$      	&96.67  &$\textbf{1.826}$      &7	&98.36    &27.132    &19	&96.53  &112.239     &132	    &$\textbf{99.44}$ &62.054 \\ \hline
\end{tabular}}
\end{table*}%(table*)??
\begin{table*}[!t] %(table*)
\centering
\caption{Experimental results of offline feature selection methods on benchmark data sets.}% by (a) Alpha-investing, (b) Fast-OSFS, (c) Baseline, and (d) OGFS.}
\label{resOff}
\centering

\begin{tabular}{|l|r | r| r| r| r |r| r| r | r | r |r|} %{|c| c| c| c| c| c |c| c |c|}
\hline
\multirow{3}*{Data Set} & \multicolumn{3}{c|}{MI} & \multicolumn{3}{c|}{LARS} & \multicolumn{3}{c|}{GBFS} & \multicolumn{2}{c|}{Baseline}\\
\cline{2-12} &$\#$dim.	&accu.	&time(s)            &$\#$dim.	&accu. &time(s) &$\#$dim.	&accu. &time(s) &$\#$dim.	&accu.\\ \hline
Wdbc     &\textbf{20} &\textbf{95.96} &0.02 &21 &95.61 &0.96 &23 &94.74 &1.09 &30 &95.26  \\ \hline
Ionosphere	& \textbf{20} &\textbf{92.61} &0.01 &32 &92.04 &0.86 &32 &91.48 &0.80 &34 & 92.05 \\ \hline
Spectf    &\textbf{20} &80.20 &0.01 &44 &80.56 &0.86 &31 &80.19 &0.81 &44 &\textbf{80.56} \\ \hline
Spambase	&\textbf{20}       &91.02     &0.20       &84     	&\textbf{94.07} &0.55 &55	&92.28 &0.76 &46	&93.09 \\ \hline
Colon     &20 &82.38 &0.33 &58 &85.95 &0.87 &\textbf{4} &\textbf{92.14} &1.05 &2000 &84.05   \\ \hline
Prostate	 &20 &92.00 &1.02 &98 &94.09 &0.97 &\textbf{5} &\textbf{96.00} &1.97 &6033 &90.00 \\ \hline
Leukemia  &20 &94.64 &1.16 &70 &\textbf{100.00} &0.94 &\textbf{3} &94.46 &1.88 &7129 &90.36 \\ \hline
Lungcancer	 &20 &99.44 &2.34 &166 &\textbf{100.00} &1.55 &\textbf{3} &97.22 &4.16 &12533 &96.11 \\ \hline
\end{tabular}
\end{table*}

\begin{table}[!t] %(table*)
\centering
\caption{Image classification results on the Cifar-10 data set with random feature groups.}
\label{cifarGroup}
\centering
\begin{tabular}{|c|l| c|r| c| }
\hline
Index & Order &$\#$dim. & accu. & time(s)\\ \hline
1	& 1 5 3 4 2 & 1,969  & 48.49 	& 137.96  \\ \hline
2	& 2 1 3 4 5	& 1,973  & 49.27	&141.18 	 \\ \hline
3	& 2 5 3 1 4 &1,992 & 48.57 & 136.62 \\ \hline
4	& 4 5 2 1 3	& 1,988 & 48.44 & 135.03 \\ \hline
5	& 3 1 4 5 2 & 1,955 & 48.64 & 137.20  \\ \hline
6	& 2 4 3 5 1 & 1,948 & 48.11 & 137.07 \\ \hline
7   & 5 3 4 1 2 & 1,995 & 48.48 & 135.55 \\ \hline
8   & 1 5 4 3 2 & 1,978 & 48.54 & 137.28 \\ \hline
9   & 3 5 4 2 1 & 1,977 & 47.92 & 138.32 \\ \hline
10  & 1 3 4 2 5 & 1,981 & 49.28 & 142.85 \\ \hline
average &- &- &48.58$\pm$0.43 & - \\ \hline
\end{tabular}
\vspace{-1em}
\end{table}

\subsection{On the Influence of Group Orders}
In this part, we show the performance of our method regarding to the order of feature groups in Table~\ref{cifarGroup}. The experiment is conducted on the Cifar-10 dataset. The original feature space is $F=[G_1,G_2,G_3,G_4,G_5]$. We randomly generated the order of the groups of features 10 times, as shown in the second column of Table~\ref{cifarGroup}. Our algorithm obtains an average accuracy of 48.58\%. The standard variation of the accuracy is 0.43. To sum up, the order of the feature groups has some influence towards the method, but the variation is within certain range. Thus, it demonstrates that our method is stable in real-world applications.

\subsection{Experimental Results on UCI Data Sets}
Table~\ref{bench} lists the eight benchmark data sets from UCI repository %\footnote{http://archive.ics.uci.edu/ml} %\cite{UCI}
(Wdbc, Ionosphere, Spectf and Spambase) or microarray domains\footnote{http://www.cs.binghamton.edu/~lyu/KDD08/data/} (Colon, Prostate, Leukemia, and Lungcancer). Note that, for these eight data sets, there is no natural group information, and the group structure is generated by randomly dividing the feature space. This experiment can help us test the robustness of the OGFS approach.

After feature selection, we test the performance of the feature space based on the three classifiers, $k$-NN, J48, and Randomforest in Spider Toolbox\footnote{http://www.kyb.mpg.de/bs/people/spider/main.html}. We adopt 10-fold cross-validation on the three classifiers and choose the average accuracy as the final result. Table~\ref{resultBench}  shows the experimental %experimental
results of classification accuracy versus compactness on the 8 %eight
UCI data sets.

\begin{itemize}
\item OGFS vs. Grafting%Three feature selection methods vs. Baseline

Though the Grafting uses the information about the global feature space, our algorithm outperforms Grafting on 6 out of the 8 data sets in terms of accuracy. On the 6 data sets, our method obtains 3$\sim$5\% gain in accuracy. More specifically, on the dataset Colon, the accuracy of Grafting is 84.26\%, %percent,
while OGFS achieves 90.47\%. %percent.
On the datasets Leukemia and Lungcancer, our algorithm achieves a fairly high accuracy (over 99.0\%). On the other two data sets Wdbc and Ionosphere, OGFS also obtains comparative accuracy, only 0.5\% lower. On the Ionosphere dataset, OGFS achieves better compactness. The results show that OGFS is able to obtain the features with discriminative capability.

 \item OGFS vs. Alpha-investing

Alpha-investing obtains better compactness than our OGFS algorithm on 7 data sets, but it performs worse in terms of accuracy. Our method outperforms Alpha-investing on the other 6 data sets. More specifically, on the dataset Colon, the accuracy of Alpha-investing is 79.76\%, while OGFS reaches up to 90.47\%. On the Wdbc and Ionosphere data sets, the two methods achieve comparable accurac. For instance, on the Ionosphere dataset, our algorithm achieves an accuracy of 91.47\% %percent
while Alpha-investing achieves an accuracy of 91.76\%. This is because the previously selected subset will never be reevaluated in Alpha-investing, which affects the selection of the later arrived features. However, in our algorithm, %we consider the correlations of the features within a group and the relationship between the groups.
 selected features will be reevaluated in the inter-group selection in each iteration. Thus, our algorithm is able to select sufficient features with discriminative power.

\item OGFS vs. OSFS

OSFS obtains better compactness on most of the data sets, but our algorithm is better than OSFS in accuracy on 6 out of the 8 data sets with a small compactness loss. More specifically, on the Ionosphere and Spambase data sets, the accuracy of our algorithm (91.47\%, 93.09\%) are slightly lower than OSFS (92.60\%, 94.07\%). But on the other data sets, our algorithm significantly outperforms OSFS. For example, on the dataset Colon, our algorithm achieves an accuracy of 90.47\% while OSFS reaches 85.95\%. On the Prostate dataset, our method (98.00\%) performs much better than OSFS (91.09\%). The reason is that OSFS only evaluates features individually rather than in groups. Meanwhile, different from OSFS, our algorithm facilitates the relationship of features within groups and the correlation between groups, which will lead to a better feature subset.
%\end{itemize}

In terms of time complexity, Alpha-investing is the fastest, except 0.15 seconds slower than our algorithm on the dataset Spambase. On the first 4 data sets, Grafting costs over 7 seconds on the Wdbc dataset, while the other algorithms accomplish the feature selection all in less than 1.0 second. When the feature space reaches thousands (Colon, Prostate and Leukemia), OGFS, Alpha-investing and Grafting methods take less than 15 seconds. OSFS takes 33.8443 seconds on the Colon dataset. This is because each time a relevant feature is added, redundancy analysis is triggered over all selected features. On the Lungcancer dataset, Alpha-investing takes less than 2.0 seconds. OSFS is only inferior to Alpha-investing with 27.13 seconds. OGFS costs about 1 minute, still faster than Grafting with 112.24 seconds. It demonstrates that simple consideration of each dimension of coming feature is efficient, like Alpha-investing. At the same time, the time complexity of other algorithms is correlated with not only the global feature space but also the selected features in previous stage. Although the reevaluation of selected features costs more time, they are more robust and achieve relatively better classification performance.

\item OGFS vs. Offline feature selection methods
Table~\ref{resOff} reports the results of offline feature selection methods and Baseline. LARS obtains the best accuracy. For instance, on the Leukemia dataset, LARS reaches 100.00\% accuracy, 5\% better than MI and GBFS. In most cases, MI and GBFS are comparative with LARS. The offline methods all obtain the compactness with less than 60 features. We can observe that OGFS is comparative with the best result of offline feature selection methods. The results demonstrate the efficacy of OGFS in general feature selection applications.

\end{itemize}

In summary, in term of classification accuracy, experimental results on UCI data sets show that our algorithm is superior than comparative online feature selection methods. OGFS achieves comparative results with the best offline performance. It implies that our method enjoys a significant improvement compared to state-of-the-art online feature selection models. %  while maintaining compactness and efficiency. %These experimental results demonstrate the robustness of our OGFS approach.
%%{\color{blue}
%%\subsection{The Parameter $\epsilon$}
%%Finally, we also test the sensitivity of the parameter $\epsilon$, which is proposed in the intra-group selection. We vary $\epsilon$ from 0.00001 to 0.1. Figure~\ref{ } demonstrate the performance curve with respect to the variation of $\epsilon$. Here we also illustrate the performance of the other three methods, i.e., ``Alpha-investing'', ``OGFS'' and ``Grafting'', for comparison. From the results we can see that the performance
%%}
\section{Conclusion}
\label{conclu}
In this paper, we investigate the online group feature selection problem and present an novel algorithm, namely OGFS. In comparison with traditional online feature selection, our proposed approach considers the situation that features arrive by groups in real-world applications. We divide online group feature selection into two stages, i.e., online intra-group and inter-group selection. Then, we design a novel criterion based on spectral analysis for intra-group selection, and introduce a sparse regression model to reduce the redundancy in inter-group selection. Extensive experimental results on image classification and face verification demonstrate that our method is suitable for real-world applications. %On the Cifar-10 dataset, we compared OGFS-Intra, OGFS-Inter and OGFS. It confirms that OGFS-Intra is efficient and can select most discriminative features. OGFS-Inter can control the compactness of the final feature space with a very slight accuracy decrease. 
We also validate the efficacy of our method on several UCI and microarray benchmark data sets.

%\ifCLASSOPTIONcaptionsoff
%  \newpage
%\fi

\clearpage
\bibliographystyle{IEEETran}
\bibliography{ijcai13}

\end{document}